\documentclass[acmlarge,screen]{acmart}

\usepackage{tabularx}
\usepackage{multirow}
\usepackage[ruled,vlined]{algorithm2e}
\usepackage{algorithmic}
\usepackage{listings}

\newcolumntype{C}{>{\centering\arraybackslash}X} %
\newcolumntype{P}[1]{>{\centering\arraybackslash}p{#1}}
\newcolumntype{M}[1]{>{\centering\arraybackslash}m{#1}}
\graphicspath{ {images/} }

\AtBeginDocument{%
  \providecommand\BibTeX{{%
    \normalfont B\kern-0.5em{\scshape i\kern-0.25em b}\kern-0.8em\TeX}}}

\usepackage{xcolor,soul}

\newcommand{\ds}[1]{}
\newcommand{\cm}[1]{} 
\newcommand{\ip}[1]{} 
\newcommand{\cit}[1]{} 

\newcommand{\rev}[1]{#1} 

\newcommand{\revtwo}[1]{#1} 

\setcopyright{rightsretained}

\begin{document}

\acmJournal{IMWUT}
\acmYear{2021} 
\acmVolume{5} \acmNumber{1} 
\acmArticle{36}
\acmMonth{3}
\acmPrice{}
\acmDOI{10.1145/3448112}

\title[\textit{SelfHAR}: Improving Human Activity Recognition through Self-training with Unlabeled Data]{SelfHAR: Improving Human Activity Recognition through Self-training with Unlabeled Data}

\author{Chi Ian Tang}
\email{cit27@cam.ac.uk}
\orcid{0000-0001-5674-2087}
\affiliation{
  \department{Department of Computer Science and Technology}
  \institution{University of Cambridge}
  \city{Cambridge}
  \country{UK}
}

\author{Ignacio Perez-Pozuelo}
\authornote{Both authors assert joint second authorship.}
\email{ip325@cam.ac.uk}
\affiliation{
  \department{Dept of Medicine}
  \institution{University of Cambridge}
  \city{Cambridge}
  \state{Cambridgeshire}
  \country{UK}
}
\affiliation{
  \institution{The Alan Turing Institute}
  \city{London}
  \country{UK}
}

\author{Dimitris Spathis}
\authornotemark[1]
\email{ds806@cam.ac.uk}
\affiliation{
  \department{Department of Computer Science and Technology}
  \institution{University of Cambridge}
  \city{Cambridge}
  \country{UK}
}

\author{Soren Brage}
\email{soren.brage@mrc-epid.cam.ac.uk}
\affiliation{
  \department{MRC Epidemiology Unit, School of Clinical Medicine}
  \institution{University of Cambridge}
  \city{Cambridge}
  \country{UK}
}

\author{Nick Wareham}
\email{nick.wareham@mrc-epid.cam.ac.uk}
\affiliation{
  \department{MRC Epidemiology Unit, School of Clinical Medicine}
  \institution{University of Cambridge}
  \city{Cambridge}
  \country{UK}
}

\author{Cecilia Mascolo}
\email{cm542@cam.ac.uk}
\affiliation{
  \department{Department of Computer Science and Technology}
  \institution{University of Cambridge}
  \city{Cambridge}
  \country{UK}
}

\renewcommand{\shortauthors}{Tang et al.}

\begin{abstract}

Machine learning and deep learning have shown great promise in mobile sensing applications, including Human Activity Recognition. However, the performance of such models in real-world settings largely depends on the availability of large \rev{datasets that captures diverse behaviors}. Recently, \rev{studies in} computer vision and natural language processing have shown that leveraging massive amounts of unlabeled data enables performance on par with state-of-the-art supervised models.

In this work, we present \textit{SelfHAR}, a \rev{semi-supervised} model that effectively learns to leverage unlabeled mobile sensing datasets \rev{to complement small labeled datasets}. Our approach combines teacher-student self-training, which distills the knowledge of unlabeled \rev{and labeled} datasets while allowing for data augmentation, and multi-task self-supervision, which learns robust signal-level representations by predicting distorted versions of the input.

We evaluated \textit{SelfHAR} on various HAR datasets and showed state-of-the-art performance over supervised and previous \rev{semi-supervised} approaches, with up to $12\%$ increase in F1 score using the same number of model parameters at inference. Furthermore, \textit{SelfHAR} is data-efficient, reaching similar performance using up to 10 times less labeled data compared to supervised approaches. Our work not only achieves state-of-the-art performance in a diverse set of HAR datasets, but also sheds light on how pre-training \rev{tasks} may affect downstream performance. 

 \end{abstract}



\maketitle

\section{Introduction}
The growing prevalence of wearable devices and mobile phones equipped with a plethora of sensors has yielded large amounts of unlabeled time-series data which could be leveraged to understand physical behaviors at an unprecedented scale. Human Activity Recognition (HAR) aims to accurately classify human physical activity and enable the detection of simple and complex behaviors in real-world settings from this type of data.
Traditionally, HAR through wearable and mobile sensing relied on sliding window segmentation and manual feature extraction, followed by a variety of supervised learning techniques to recognize simple and complex activities like walking, running, cycling and cooking. While in simple scenarios hand-crafted features may suffice, deep learning methods have proven to be more effective in complex HAR tasks~\cite{deep_har_1d, deep_har_2,  multi_self_har, deep_har_1,deep_har_survey_2}. Indeed, deep learning has shown promise in HAR by automatically extracting useful features for the target task~\cite{deep_feature_1, deep_feature_2}. However, the performance of both traditional HAR methods and deep learning models in real-world settings could be adversely affected by the bias and limitations introduced by conventional laboratory-based HAR datasets. It is particularly important to consider limitations due to their size, diversity and ability to capture and represent the richness, noise and complexity that can be found in free-living, unconstrained data~\cite{deep_bias, dataset_bias_1, dataset_bias_2}. These limitations on mobile sensing datasets are only exacerbated by the difficulty in collecting labeled data outside of laboratory settings~\cite{multi_self_har}. 

In computer vision and natural language processing, labeling data after it has been collected is plausible, as these data streams can be readily interpreted. However, in mobile and wearable devices, sensor data streams are abstract and not easy to interpret, often making post-hoc labeling futile. Thus, studies~\cite{hhar, motionsense, unimib, uci_har, wisdm} which collect HAR data are usually small, mostly involving less than 50 participants. Similarly, these studies follow a fixed experimental protocol and require participants to perform certain activities over a short period of time (usually under an hour). This inherent difficulty in the collection of mobile sensing data limits the size, diversity and quality of the resulting datasets. 

In order to overcome the inherent limitations of labeled datasets, semi-supervised learning techniques have been proposed to complement instances where limited labeled data are available, through the use of unlabeled data~\cite{semi_supervised_survey, semi_supervised_intro}. The main aim of these methods is to increase the diversity and quantity of data that machine learning models are trained on. %
There are many different families of semi-supervised learning techniques, with varying assumptions on the labeled and unlabeled data. Transfer learning and self-supervised learning are two of the most commonly used semi-supervised techniques, which are comparatively straightforward to implement due to their supervised nature of training.

Network-based transfer learning is a commonly used method that involves models being pre-trained on a large dataset and then fine-tuned to a different task or domain. Previous work in computer vision has proposed methods to improve the performance of object and action classification tasks with limited data by transferring a convolutional neural network trained on the large-scale ImageNet dataset~\cite{imagenet,transfer_learning_cnn}. The transfer pipeline involves freezing the pre-trained layers of the model, with only the rest of the model being updated during the training with the target dataset. This is one of the most commonly used techniques in network-based transfer learning, with an aim to combat Catastrophic Forgetting~\cite{catastrophic_forgetting, transfer_learning_frozen}, where neural networks tend to `forget' previously learned knowledge during re-training. Maxime et al. showed that models trained with a network-based transfer learning pipeline achieved state-of-the-art results on small benchmark datasets~\cite{transfer_learning_cnn}. %

Self-supervised learning is a representation learning paradigm that leverages the intrinsic structure present in the input signals and does not require pre-labeled data~\cite{lee2017unsupervised}. Self-supervised models make use of large-scale unlabeled data through specialized learning objectives in order to obtain supervision from the data \textit{itself}, using supervised loss functions. Through this process, objectives are computed from the signals themselves by applying known transformations to the input data. Most importantly, the intermediate representations capture semantic and structural meaning that can then be exploited for a variety of downstream tasks. This new paradigm has been successfully applied in filling the blanks for image datasets~\cite{iizuka2017globally}, next word prediction~\cite{bert}, video frame order validation~\cite{video_shuffle}, and, more recently, small-scale activity data~\cite{multi_self_har}.

Motivated by the success of the aforementioned semi-supervised techniques in other disciplines~\cite{image_billion, image_multi_task, multi_self_har}, our study aims to improve the performance of HAR systems when training with limited data. Saeed et al. showed that applying self-supervised learning in HAR resulted in performance improvements~\cite{multi_self_har}, however, the power of training with an unlabeled, complementary dataset was not leveraged.

\begin{figure}
    \begin{center}
        \includegraphics[trim=0 20 45 0,clip,width=\textwidth]{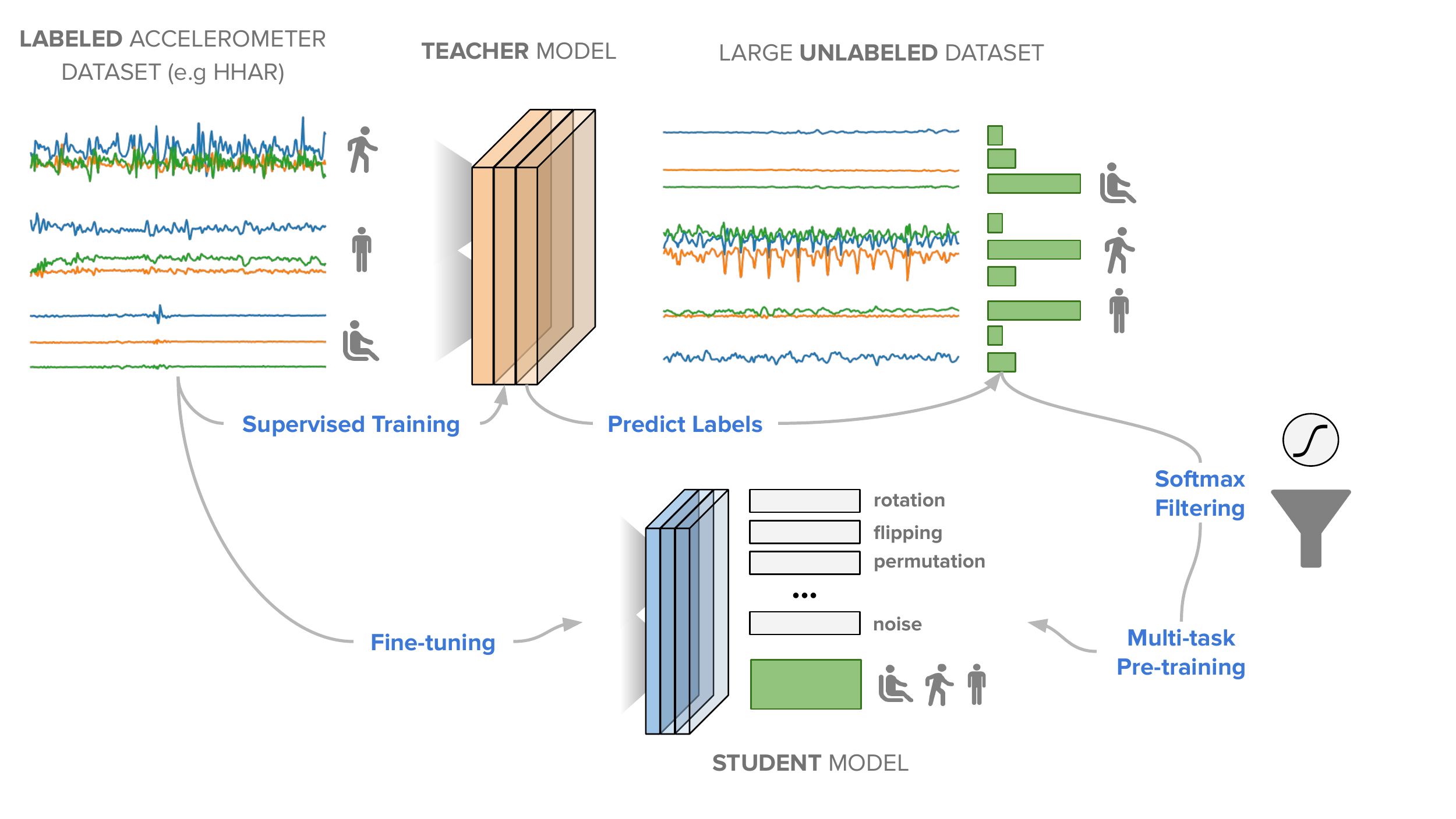}
    \caption
    {Overview of the proposed approach \textit{SelfHAR}. A teacher model distills the knowledge of labeled accelerometer data and then is used to label a large unlabeled dataset. We select only the high-confidence data-points from the previous step and train a student model to discriminate signal transformations as well as the activities. Lastly, the ground truth labels from the training set are used to fine-tune the student model.}
    \label{figure:architecture_overview}
    \end{center}
\end{figure}

We introduce \textit{SelfHAR} (see Figure \ref{figure:architecture_overview}), a semi-supervised framework that \rev{complements supervised training on labeled data with} self-supervised pre-training to leverage the information captured by large-scale unlabeled datasets in HAR tasks. This approach implicitly introduces more diverse sensor data to our training paradigm, therefore offering a unique new avenue for performance improvement. Specifically, the approach combines self-training and signal transformation discrimination as self-supervised learning tasks to increase the internal and external diversity of the training data, which allows the model to learn more generalizable features. \rev{The labeled datasets are effectively extended with the unlabeled data in this approach.} We examine the performance of our method on several datasets collected from a diverse group of participants, utilizing different devices and sensors and with different experimental setups. We demonstrate increased performance without increasing model complexity at inference time, which could have important implications for real-world deployment of machine learning models on resource-constrained mobile devices. Additionally, we evaluated the models in scenarios with limited training data, showing similar performance achieved with 10 times less labeled data compared to \rev{the} conventional fully supervised approach. %

Our work makes three major contributions:
\begin{enumerate} 
    \item We introduce \textit{SelfHAR}, a training paradigm \rev{that incorporates the teacher-student setup into self-supervised pre-training. This combination of methods leverages large unlabeled wearable and mobile sensing datasets more effectively, enabling deep learning models to learn feature extraction and representations in a multi-task setup.}
    
    \item We evaluate \textit{SelfHAR} on seven different datasets, consisting of different sensor types, populations and protocols. We demonstrate that leveraging unlabeled datasets to increase the device, placement and user diversity leads to a further improvement in performance, outperforming state-of-the-art supervised and semi-supervised methods by up to $12\%$. This superior performance was attained without increasing the complexity of the machine learning model. \rev{With our models, we show that solely using pre-training can only go so far in terms of handling diverse unlabeled datasets.}
    
    \item Our proposed approach shows robust results in cases where limited amounts of data are available. An F1 score of $0.67 - 0.88$ was attained with only 10 samples per class, and a similar performance was achieved with up to 10 times less labeled data compared to \rev{the} conventional fully supervised approach. %
\end{enumerate}

\section{Related Work and Motivation}
In this section, we provide an overview of research related to our work, highlighting the gaps which this work aims to fill.

\subsection{Human Activity Recognition}
Human activity recognition involves identifying actions performed by a person based on data of the subjects and the surroundings. Usually, the set of actions to classify reflect common movements performed by people, such as standing, walking, running or cycling, while the data can pertain to different types of sources and sensors. HAR systems can be broadly categorized into two main types: sensor-based HAR and video-based HAR~\cite{har_review_privacy}. 

In recent years, the use of deep learning for human activity recognition has become more prevalent~\cite{guan2017ensembles, deep_har_context, peng2018aroma, krishna2018lstm}. Convolutional neural networks have been commonly used~\cite{deep_har_cnn_early, deep_har_1d, deep_har_2, peng2018aroma,zhai2020making} due to their ability to capture spatial relationships between sensor signals. Yang et al. demonstrated strong performance by using neural networks with temporal convolutions (1D convolution) for automatically extracting useful features from raw sensor signals to be used in human activity recognition tasks~\cite{deep_har_3}. Subsequent works~\cite{ deep_har_1d, deep_har_2, deep_har_1} have further proved the effectiveness of deep neural networks in developing accurate HAR systems.

However, although deep learning has been effective in extracting useful features in laboratory datasets, the real-world performance is dependent on the size, the diversity and the ability of the data to represent the real characteristics of the actions~\cite{deep_bias, dataset_bias_1, dataset_bias_2}. This problem is amplified in HAR due to the inherent difficulty in the collection of labeled sensor data.

\subsection{Semi-supervised Learning} \label{subsection:related_work_semi_supervised}
Autoencoders~\cite{image_autoencoder} and generative methods~\cite{sensegan, deep_semi_generative} are some of the new methods aiming to mitigate some of the inherent limitations associated with supervised learning techniques. These are examples of semi-supervised learning, %
which is a broad range of methods that aim to utilize unlabeled data to complement and circumvent the limitations of using labeled data by improving upon the quantity and diversity of data used for training~\cite{semi_supervised_survey, semi_supervised_survey2, semi_supervised_intro, semi_har}.

\rev{
In human activity recognition, due to the difficulty in collecting labeled sensor data, semi-supervised learning has been actively researched {\cite{en_co_training_guan2007activity, sparse_coding_bhattacharya2014using, multi_self_har}}. The {\textit{En-Co-training}} {\cite{en_co_training_guan2007activity}} algorithm combines the co-training and ensemble learning paradigms to leverage unlabeled sensor data. The algorithm iteratively trains an ensemble of several different classifiers (an ensemble of a decision tree, a Na\"ive-Bayes classifier and a k-nearest neighbor classifier was suggested by the authors), and uses the trained classifiers to predict the labels for a pool of unlabeled samples. The samples on which all classifiers agree are put into the training set. After a predefined number of iterations, the final predictions are obtained by majority voting from the ensemble. Although the paradigm can incorporate unlabeled data, the iterative approach requires training different many models from scratch, and this becomes costly when the amount of unlabeled and labeled data is large. Also, the diversity requirement limits the choice of classifiers, where they need to be sufficiently different from each other to provide a proper supervisory signal.

In another work, Bhattacharya et al. proposed a sparse-coding framework for using unlabeled data in HAR {\cite{sparse_coding_bhattacharya2014using}}. The framework leverages unlabeled data by deriving a sparse-coding dictionary of basis vectors from them for linear reconstructions of signals. The basis vectors are filtered by their empirical entropy, and then they are used to decompose new samples, giving linear coefficients (activation) that are used as features. A classifier, such as a support vector machine, is then trained on these features for HAR. State-of-the-art results were reported in the study, but the requirement of solving L1-regularized least square problems during dictionary learning and prediction makes it difficult to scale to a large amount of data.
}

\subsubsection{Self-supervised Learning} 
Within semi-supervised training methods, self-supervised learning, in which artificial tasks are designed to generate labels for unlabeled data, has been actively studied due to their simplistic design, high similarity to supervised methods and the ability to exploit the invariants and properties of the data itself to provide a supervisory signal. 

In the computer vision community, many different tasks have been designed to train models with a limited amount of data, for instance: colorization, where the model is trained to predict the actual colors from images which were turned into gray-scale~\cite{image_colourisation}; jigsaw puzzle-solving, where images are divided into patches and the model is trained to determine the relative positions~\cite{image_jigsaw}; or autoencoders, where the model is trained to reconstruct images~\cite{image_autoencoder}. Similarly, self-supervision has also been applied to natural language processing (NLP). The Bidirectional Encoder Representations from Transformers (BERT)~\cite{bert} achieved state-of-the-art performance by pre-training a model to predict different occluded parts of a sentence. The success of self-supervised learning techniques in computer vision and NLP indicates their potential in improving deep learning models, including those for HAR, but adaptations are necessary to handle the differences in modalities in data. 

In human activity recognition, Saeed et al. introduced the concept of signal transformation discrimination as a self-supervised task to capture invariant representations in mobile sensor signals~\cite{multi_self_har}. In their study, they selected eight signal transformations as the source of the supervisory signal. The signal transformation discrimination task requires models to learn whether a sensor data sample has been transformed. Compared to the purely supervised training approach, they reported a performance gain by pre-training the model with the transformation discrimination task, using only the target dataset for pre-training in most datasets they tested. %

Although Saeed et al. briefly explored the possibility of using another unlabeled dataset for pre-training, their architecture yielded little to no improvement when compared to using the target dataset for pre-training. 
These results suggested there was potential room for improvement when using unlabeled datasets through the design of new architectures that truly leverage the information carried in these unlabeled data.

\revtwo{
Sarkar et al. adapted the transformation discrimination task to electrocardiogram (ECG) data for emotion recognition \cite{ecg_transformation_sarkar2020self}. Overall, the proposed approach is very similar to that in Saeed et al. \cite{multi_self_har}, where the authors selected a similar set of transformation functions for signal transformation discrimination pre-training, followed by fine-tuning the models on the target dataset. A deeper investigation into the impact on the recognition accuracy caused by changing the difficulty of the transformation tasks and using different sets of unlabeled data, as well as the relationship between downstream (target) and upstream (pre-training) tasks, was performed. The authors reported state-of-the-art results on different datasets compared to previously proposed approaches. This work gave insights into the relationship between the difficulty of pre-training tasks and the accuracy of the models, but whether these results can be applied to HAR is yet to be studied.

As a generalization of the aforementioned work, Saeed et al. proposed the \textit{Sense and Learn} framework which leverages self-supervision for representation learning on multi-modal sensor data \cite{sense_and_learn_saeed2020sense}. The authors proposed a set of eight separate self-supervised tasks, from which practitioners can choose according to their knowledge on the modality of data and the target tasks, or the empirical results. The set of self-supervised tasks included ones that leverage multi-modal sensor data, such as \textit{Blend Detection}, which detects whether the sensor signals from different sources have been blended, ones which detect changes in the data, such as transformation discrimination and \textit{Odd Segment Recognition}. Although a high performance was reported when the number of labeled samples is particularly limited, the fully-supervised models often out-performed the best-performing self-supervised methods. Furthermore, some of the self-supervised tasks rely on multi-modal data, which might be expensive to collect in terms of system resources, especially in real-time.

Spathis et al. utilized the underlying physiological relationship between the heart rate responses to movement as a supervisory signal for the extraction of user-level physiological embeddings \cite{self_supervised_physiological_spathis2020self}. The authors pre-trained a network to predict heart rate signals based on accelerometer data and contextual metadata such as the hour of the day. Apart from being able to use the trained model as a heart rate estimator, the representations extracted by the model can be aggregated for each user to form user-level embeddings, which was shown to out-perform other methods in predicting fitness and demographic variables, such as height and BMI. This study demonstrated the ability of self-supervised models in extracting useful representations for a wide range of tasks towards leverage multi-modal data. However, it does not explicitly learn from both unlabeled and labeled data, as well as it is not clear whether such model could also help the task of HAR.
}

\subsubsection{Self-training}
Self-training is another semi-supervised training technique that involves an iterative process of training that can leverage large-scale unlabeled data. The basic set up of a self-training pipeline consists of (1) training a teacher model with the labeled data, (2) using that teacher model to obtain labels for the unlabeled data, and (3) training a student model on both the original labeled data and the teacher-labeled data~\cite{semi_supervised_survey, semi_supervised_survey2}. This approach has been used in several contexts, for instance, for word sense disambiguation~\cite{yarowsky1995unsupervised}, %
in object detection~\cite{self_training_object_early}, as well as in image classification~\cite{image_billion} and semantic segmentation~\cite{self_training_semantic}.

Yalniz et al. proposed a method to leverage billions of unlabeled images for image classification by self-training~\cite{image_billion}. Their proposed method consisted of a simple self-training strategy: (1) train a teacher model on the target labeled dataset; (2) generate self-training labels for the unlabeled images using the teacher model; (3) for each class, select the most confident predictions by the teacher model and combine them to form a self-training dataset; (4) pre-train a student model with this dataset and finally (5) fine-tune the student model with the target labeled dataset. This self-training setup was shown to be effective in improving the accuracy of a range of deep learning models in different tasks, including image classification and video classification, %
achieving state-of-the-art performance when published. Overall, this study showcased an effective way to utilize unlabeled datasets which we used as inspiration for our own architecture. However, this work is reliant on millions of labeled images available in the ImageNet dataset and very deep neural network architectures. The ability to translate such improvement in performance to mobile sensing is yet to be studied.

\subsection{Transfer Learning}
Transfer learning is another commonly used method to mitigate the need for collecting a large labeled dataset for specific tasks. Transfer learning methods are based on the assumption that models could learn to solve new problems faster or better by applying knowledge that was previously learned~\cite{transfer_learning_survey, transfer_learning_survey_2}. %
Tan et al. categorized transfer learning techniques in deep learning into four categories: (1) instances-based, which involves selecting samples that are similar to the target dataset, (2) mapping-based, which aims to map data from both datasets to a common feature space, (3) network-based, which re-uses parts of another model trained with the other dataset, and (4) adversarial-based, which involves adversarial training in making the models learn to extract domain-independent features~\cite{transfer_learning_survey_2}.

Network-based transfer learning is a commonly used method which involves models being pre-trained on a large dataset and then fine-tuned to a different task or domain~\cite{transfer_learning_cnn, transfer_learning_cnn_2}. Maxime et al.~\cite{transfer_learning_cnn} proposed a method to improve the performance of object and action classification tasks with limited data by transferring a convolutional neural network trained on the large-scale ImageNet dataset~\cite{imagenet}. The transfer pipeline involves freezing the pre-trained layers of the model, with only the rest of the model being updated during the training with the target dataset. This is one of the most commonly used techniques in network-based transfer learning which aims to overcome \textit{catastrophic forgetting}~\cite{catastrophic_forgetting, transfer_learning_frozen}, where neural networks tend to `forget' previously learned knowledge during re-training. Maxime et al. showed that models trained with their proposed pipeline achieved state-of-the-art performance on benchmark datasets with limited data~\cite{transfer_learning_cnn}. These works present an effective way to balance the knowledge learned from different datasets in different training stages for a deep learning model.

\subsection{Motivation}
The above literature highlights a clear gap in utilizing large-scale unlabeled data in a self-training setting and limited success in leveraging unlabeled datasets using self-supervised learning in HAR. In this work, we propose a training strategy that incorporates self-training and self-supervised learning for HAR to effectively utilize unlabeled data by learning the semantic and structural meaning within sensor signals. Our strategy also incorporates techniques from transfer learning, where parts of the network are frozen during fine-tuning. This is to ensure that the knowledge of general feature extraction is retained while the model is fine-tuned to the final task. To the best of our knowledge, our study is the first to utilize self-training in a deep learning setup for HAR and demonstrate that it is effective in leveraging large and unlabeled sensor data. We describe the details of our approach next. %

\section{Approach}
In this section, we introduce \textit{SelfHAR}, our semi-supervised training framework for HAR, which incorporates self-training and self-supervised learning for leveraging unlabeled datasets to improve the diversity of data that the models are trained on. %
\subsection{Problem Statement}
We consider the problem of HAR, defined as follows (inspired by the definition provided by Lara et al.~\cite{larasurvey}):
given a set \(D\) of time series data, a set \(A = \{a_i, \ldots, a_n\}\) of \(n\) activity labels, and a set \(W = \{W_1, \ldots, W_m\}\) of \(m\) time windows defining start and end-points of time periods, the task is to find a mapping function \(f\) such that \(f(D, W_i)\) outputs an activity label \(a_i\), where \(a_i\) should represent the actual activity performed by the subject during \(W_i\). In addition to the labeled dataset \(D\), we consider the scenarios where there is a complementary dataset \(U\), collected from similar sensors as \(D\) but without labels. We aim to train models such that improvement in recognition accuracy can be achieved by using \(D \cup U\) over using only \(D\) (i.e., by leveraging an unlabeled dataset). The evaluation is conducted by asking the models to predict the activity labels of an unseen subset of \(D\).

\subsection{The Training Pipeline} \label{section:training_pipeline}

\subsubsection{Overview}

\begin{figure}
    \begin{center}
      \includegraphics[width=\textwidth]{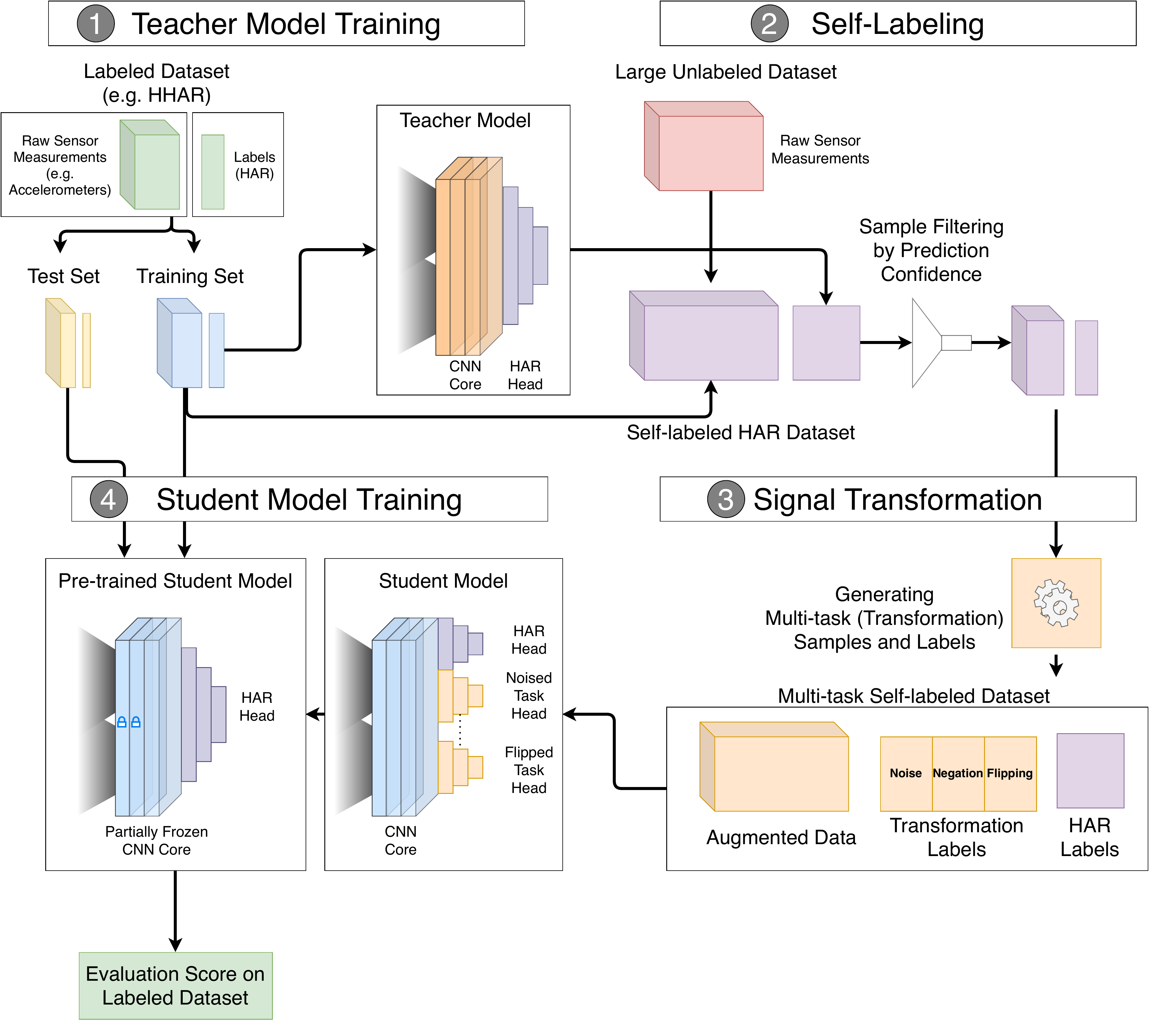}
    \caption{Detailed schematic of the proposed pipeline \textit{SelfHAR}. The pipeline adopts a teacher-student setup, where a teacher model is first trained and distills the knowledge of labeled data by labeling a large unlabeled dataset. High-confidence data-points from the previous step are selected and then augmented through signal transformation to form a multi-task training dataset. The student model is first pre-trained to discriminate signal transformations as well as the activities, and the ground truth labels from the training set are then used to fine-tune it. Evaluation is then performed on an unseen subset of the target dataset.}
    \label{figure:architecture_flow}
    \end{center}
\end{figure}

The proposed training pipeline combines self-training and self-supervised pre-training into a multi-step training pipeline (see Figure~\ref{figure:architecture_flow}):
\begin{itemize}
    \item [1.] We employ a knowledge distillation paradigm~\cite{hinton2015distilling} where a teacher model is first trained on the labeled dataset \(D\).
    \item [2.] The labels from the labeled dataset \(D\) are removed and mixed with the labeled dataset \(U\), forming a mixed dataset \(W\) without labels.
    \item [3.] The mixed dataset \(W\) is then labeled by running the teacher model, where these labeled samples are filtered by a minimum confidence threshold \(C\) and the top \(K\) samples for each class is selected, forming an intermediate dataset \(S\).
    \item [4.] The selected samples \(S\) are augmented by the eight signal transformations used in~\cite{multi_self_har}: adding random noise, scaling by a random scalar, applying a random 3D rotation, inverting the signals, reversing the direction of time, randomly scrambling sections of the signal, stretching and warping the time-series, and shuffling the different channels. This generates augmented sensor data and transformation labels, forming the self-supervised dataset \(D'\).
    \item [5.] A student model is pre-trained with the self-supervised dataset \(D'\) in a multi-task learning setting.
    \item [6.] The student model is then fine-tuned on the initial labeled dataset \(D\) and evaluated.
\end{itemize}

This pipeline is designed to increase both the internal and external diversity of the training data, which better leverage the unlabeled data and allows the model to learn more generalizable features. \revtwo{The algorithm of the training pipeline is given in Appendix~\ref{appendix:algorithm}.}

We describe the individual steps in our training pipeline in detail in the following sections.

\subsubsection{Teacher Model Training}
First, we employ a knowledge distillation paradigm~\cite{hinton2015distilling} where a teacher model is trained on the labeled dataset \(D\) following conventional deep learning training: a multi-class activity classification loss function is used with gradient descent and regularization to train the teacher model iteratively. The main goal of this step is to train a model to generate activity labels for self-training. The cross-entropy loss for multi-class classification is used for training and can be defined as:

\begin{equation}
    L_\mathrm{classification} = - \frac{1}{|D|} \sum_{d\in D} \sum_{a \in A}y_{d,a} \mathrm{log} (\{M_{\theta}(d)\}_a)
\end{equation}

Where $D$ is the labeled dataset, $A$ is the set of activity labels (activity classes), $M_{\theta}$ is the model parameterized by $\theta$, $y_{d,a}$ is $1$ if the ground-truth activity label is $a$ in window $d$ and $0$ otherwise, and \(\{M_{\theta}(d)\}_a\) is the probability of window $d$ having label $a$ predicted by the model.

\subsubsection{Samples Labeling and Selection}
After the teacher model is trained, activity labels for the mixed dataset \(W\) are generated by running the teacher model on the sensor signals. For each activity class, samples are ranked by the confidence in that particular class, and the top \(K\) samples out of those which have a softmax score of at least \(C\) in that class are selected. The probabilistic labels are kept during selection and used for training the student model. This step is to select the most confident samples in order to limit the labeling noise in each class~\cite{image_billion} while curating a large labeled dataset. The confidence score threshold is used to make sure samples that the teacher model is uncertain about do not get selected.

\subsubsection{Data Augmentation and Further Labeling}
Once the samples with the highest softmax scores are selected to form \(S\), eight signal transformation functions, as used in~\cite{har_transformations} and~\cite{multi_self_har}, are used to augment the dataset: adding random noise, scaling using a random scalar, applying a random 3D rotation, inverting the signals, reversing the direction of time, randomly scrambling sections of the signal, stretching and warping the time-series, and shuffling the different channels. \rev{These tasks are selected because mobile devices can be put in very different environments during use. For example, smartphones can be placed in a pocket, a bag, or held in a user's hand while performing different activities. Variations among devices and sensors are also common, which causes the signal to exhibit different forms of noises, including a change in orientation or signal magnitude. It is our goal to enable the models to be resilient against such noises and to learn invariances within the signals.}

Using the transformation functions, each sample in the dataset \(S\) is augmented to a total of nine different versions consisting of the original and one for each transformation. These augmented samples come with eight binary transformation labels, each indicating whether a particular transformation has been applied. There is not any mixing of two or more transformations applied to a sample simultaneously to avoid confusion and for ease of evaluation. The HAR activity labels are duplicated from the original signal, forming a multi-task self-training dataset \(D'\).

\subsubsection{Student Model Training}
The student model is a multi-task prediction model with nine prediction tasks: eight binary transformation discrimination tasks and one multi-class activity classification. It is trained in a supervised manner using the dataset \(D'\). A combined loss function is formed by summing individual losses for each task, allowing the model to learn all tasks simultaneously. After training, the early layers of the student model are fixed (frozen), and the activity classification branch is extracted and fine-tuned using the original dataset \(D\). This is similar to transfer learning setups where a part of the network is frozen and the common layers are transferred to a new model. Through this process, the models are aligned to the target HAR task, while useful feature extraction layers early in the model are retained. Early stopping is also used in both pre-training and fine-tuning to reduce over-fitting. 

The classification loss for the HAR task, the classification loss for the transformation discrimination task, and the combined loss function are as follows: %

\begin{equation}
    L_\mathrm{classification}^{\mathrm{HAR}} = - \frac{1}{|D'|} \sum_{d\in D'} \sum_{a \in A}y_{d,a}^{\mathrm{HAR}} \mathrm{log} (\{M_{\theta'}(d)\}_a^{\mathrm{HAR}})
\end{equation}

\begin{equation}
    L_\mathrm{classification}^{\mathrm{TD}} = \sum_{t \in T} \left[ -\frac{1}{|D'|} \sum_{d\in D'} \sum_{a \in \{\mathrm{True}, \mathrm{False}\}}y_{d,a,t}^{\mathrm{TD}} \mathrm{log} (\{M_{\theta'}(d)\}_{a,t}^{\mathrm{TD}}) \right]
\end{equation}

\begin{equation}
    L_\mathrm{total} = L_\mathrm{classification}^{\mathrm{HAR}} + L_\mathrm{classification}^{\mathrm{TD}} + \beta(||\theta'||^2)
\end{equation}

Where $L_\mathrm{classification}^{\mathrm{HAR}}$ represents the loss function for the HAR task, $L_\mathrm{classification}^{\mathrm{TD}}$ is the loss function for the transformation discrimination task, $D'$ is the combined self-supervised training dataset, $T$ is the set of transformation functions, \(\{M_{\theta'}(d)\}_a\) is the confidence of window $d$ having label $a$ predicted by the student model $M_{\theta'}$, $y_{d,a,t}^{\mathrm{TD}}$ is $1$ if the signal $d$ was transformed from the original using the function $t$ and $0$ otherwise. $\beta$ is the coefficient of the L2 regularization which is applied to the model parameters $\theta'$.

\rev{
\subsection{Configurations of the SelfHAR Pipeline} \label{subsection:expanded_training_pipeline}

From Figure \ref{figure:architecture_flow} and the descriptions in Section \ref{section:training_pipeline}, the \textit{SelfHAR} training pipeline can be decomposed into 4 components: (1) Teacher Model Training, (2) Self-Labeling, (3) Signal Transformation and (4) Student Model Training. In addition to this configuration, the components can be arranged to form other pipelines. In particular, the transformation discrimination task can be used on its own for self-supervised training of the teacher model, instead of combined into the student training setup (component 3). As using the transformation discrimination task for pre-training the teacher is performed before fine-tuning the teacher (component 1), it will be denoted component 0 in our study (see Appendix \ref{appendix:configurations_of_selfhar} for a visualization of the components).

A total of five different configurations were explored in this study: (1) fully supervised (using component 1 only), (2) transformation discrimination training (using components 0 and 1), (3) self-training (using components 1, 2 and 4), (4) transformation knowledge distillation (using components 0, 1, 2 and 4) and (5) \textit{SelfHAR}, which follows our approach proposed above (using components 1, 2, 3 and 4).

The fully supervised training pipeline follows conventional supervised training, in which only one model is trained through supervised learning. The transformation discrimination pipeline pre-trains models with the transformation discrimination task using the unlabeled dataset. After pre-training, the convolutional core was transferred and a randomly initialized activity recognition head was attached to the convolutional core. The final layer of the convolutional core and the new activity recognition layers were then fine-tuned on the labeled dataset. This follows closely to the method proposed by Saeed et al. \cite{multi_self_har}.

On the other hand, the self-training pipeline consists of a pure teacher-student setup without signal transformation. The pipeline follows closely the \textit{SelfHAR} setup, but the self-labeled HAR dataset was not augmented with signal transformation and the student model was pre-trained purely on the HAR labels generated by the teacher model. The student was similarly fine-tuned with the labeled dataset at the end and evaluated with the same test set.
The transformation knowledge distillation pipeline enhances the self-training only pipeline by pre-training the teacher model with the task of transformation discrimination. 

It is important to note that these models share the same neural network architecture, and the resulting prediction models share the same complexity. An ablation study is performed to evaluate these different pipelines in this study (see section \ref{subsection:results_ablation}).
} %
\section{Experimental Protocols}
In this section, we describe the datasets that we used to evaluate our proposed approach as well as the associated experimental protocols.

\subsection{Datasets}

\begin{table}
    \caption {An overview of datasets used in this project. A total of eight datasets were used in this study with different sizes, number of users, activities and device placements.}
    
    \label{table:dataset_overview}
    {\small
    \begin{center}
        \begin{tabularx}{\textwidth}{C|C|C|C|C}
            Dataset & Users & Activity Classes & Data Samples Used & Device Placement \\
            \hline
            HHAR  & 9     & 6     & 56383 & Waist and Arm \\
            \hline
            MotionSense & 24    & 6     & 6630  & Front Pocket (Trousers) \\
            \hline
            MobiAct & 66    & 11    & 57995 & Front Pocket (Trousers) \\
            \hline
            PAAS  & 28    & 11    & 32558 & Wrist \\
            \hline
            UniMiB SHAR & 30    & 9     & 1551  & Front Pocket (Trousers) \\
            \hline
            UCI HAR & 30    & 6     & 2543  & Waist \\
            \hline
            WISDM & 29    & 6     & 5435  & Front Pocket (Trousers) \\
            \hline
            Fenland (Unlabeled) & 2096  & N/A   & 168687 (Subset) & Wrist \\
        \end{tabularx}
    \end{center}
    }
\end{table}

Eight datasets were used in this study, where seven of them are labeled and one of them is unlabeled (see Table \ref{table:dataset_overview}). %
A description of each dataset used in our work is provided below.

\subsubsection{HHAR} The Heterogeneity Activity Recognition dataset~\cite{hhar} is a dataset collected to investigate the impact of heterogeneous devices on activity recognition performance, with a focus on the variations between different sensors, devices and workloads.
The data was collected from 9 participants (aged between 25 and 30) who performed 6 common daily activities: biking, sitting, standing, walking, walking upstairs and walking downstairs. The data from accelerometers and gyroscopes embedded in 8 smartphones placed around the users' waist and 4 smartwatches around the users' arms with varying sampling frequencies were collected. Two of each of the following devices were used: LG Nexus 4 (200 Hz), Samsung Galaxy S Plus (50 Hz), Samsung Galaxy S3 (150 Hz), Samsung Galaxy S3 mini (100 Hz), LG G (200 Hz) and Samsung Galaxy Wear (100 Hz). The participants were asked to perform the 6 activities for 5 minutes each, to ensure an equal data distribution among the different classes~\cite{hhar}. %
Only the sensor signals from the triaxial accelerometer of the smartphones were used in this study due to the need for compatibility among datasets. %

\subsubsection{MotionSense} The MotionSense~\cite{motionsense} dataset was collected for the development of privacy-preserving sensor data transmission systems, where the data was used to test whether personal attribute fingerprints could be extracted from the data.
24 subjects (14 men and 10 women) were recruited. The body mass of the subjects ranged between 48 kg and 102 kg, their height ranged between 161 cm and 190 cm, and their age ranged between 18 and 46. The acceleration, gravity and attitude (pitch, roll, yaw) were collected at 50 Hz from an iPhone 6s placed in the trousers' front pocket, while performing 6 activities: walking downstairs, walking upstairs, walking, jogging, sitting, and standing. 15 trials were conducted in the same environment and condition around the Queen Mary University of London's Mile End campus, with each trial lasting between 30 seconds and 3 minutes.
In our study, only signals from the accelerometer embedded in the iPhone 6s were used. %

\subsubsection{MobiAct} The MobiAct~\cite{mobiact} dataset (second release) consists of accelerometer, gyroscope and orientation readings from a Samsung Galaxy S3 (LSM330DLC inertial module at 20 Hz) for the purpose of developing a combined system which can perform HAR and fall detection. 
66 participants (51 men and 15 women), of age 20-47, height 160-193 cm and body mass 50-120kg, were recruited to perform 12 types of activities of daily living (ADLs) and 4 types of falls. Fall-related activities, including sitting, stepping in a car and jumping, were selected in order to test the robustness of the developed system. A total of 3200 trials were conducted with the participants performing the ADLs and acting out 5 different daily living scenarios. The smartphone was placed freely in the participant's pockets without specifying orientation, %
with the exception of fall detection where the smartphone was placed in the pocket on the opposite side of where the participant would be landing.  

For our work, only data unrelated to falls was used as we were interested in activity classification. These include 11 types of activities: standing, walking, jogging, jumping, walking upstairs, walking downstairs, standing to sitting on a chair (activity transition), sitting on a chair, sitting to standing, stepping into a car and stepping out of a car.

\subsubsection{PAAS} The Physical Activity Annotation Study (PAAS) was established to develop algorithms for physical activity type classification based on a single sensor location while developing a multi-sensor system to serve as a gold standard for physical activity classification in future studies. A detailed description of the study, the sensors used and the activity breakdown has been previously published~\cite{paas}.
The study recruited 28 healthy adult participants (15 women, 13 men) who were tested at the Institute of Metabolic Science at Addenbrookes Hospital in Cambridge, UK. %
For the PAAS study, participants wore nine triaxial raw accelerometers (GENEA). Additionally, four other sensors were positioned on the participant's body, including one Actigraph GT3X (Actigraph), one Sensewear (Bodymedia), and two ActiWare Cardio monitors (CamNtech). None of the monitors was obstructive of normal body movement. %
All participants performed a protocol that included 60 activities, which aimed to capture the majority of activities that an individual who works at an office does during a normal day. An audio recording was used to provide participants with activity instructions for the different sections of the protocol, ensuring that all participants' activities were recorded for the same amount of time. The full activity breakdown has also been described previously in detail~\cite{paas}. The protocol aimed to be representative of occupational activity and capture the sedentary behaviors associated with most desk jobs. Participants also wore a subset of the sensors in free-living conditions after the protocol, but this data won't be utilized for our experiments.

For our analysis, due to the relatively small size of the dataset and for simplicity purposes, activities were grouped into 11 categories: sitting, standing, walking, walking upstairs, walking downstairs, lying, cycling, home activities, office activities, personal care and shopping. In our experiments, we only used the triaxial accelerometer (GENEA, previously described) placed on the non-dominant wrist of every participant. This sensor has a dynamic range of $\pm 6g$ and sampled at 80Hz. Real-time stamps were stored for the full recording.   

\subsubsection{UniMiB SHAR} The UniMiB SHAR~\cite{unimib} dataset was also collected for HAR and fall detection.
The data were collected from 30 healthy adults (24 women and 6 men), of age 18-60, height 160-190 cm and body mass 50-82 kg. They were asked to perform 9 types of ADLs and 8 types of falls, which were selected given their popularity in other public datasets. Acceleration data from a Samsung Galaxy Nexus I9250 (with Bosh BMA220 acceleration sensor) placed in the front trouser pockets were collected at 50 Hz, and audio recordings were also collected for data annotation. 
Four different sequences were designed to allow the participants to perform with ease. Among the trials, the smartphones were placed in the left trouser pocket for half of the time, and in the right one for the other half of the time. Written informed consent was obtained from the participants, in accordance with the World Medical Association (WMA) Declaration of Helsinki.
Data for the 9 classes of ADLs were used in this study: standing up from laying, lying down to standing, standing up from sitting, running, sitting down, going downstairs, going upstairs, walking, and jumping.

\subsubsection{UCI HAR} The UCI HAR~\cite{uci_har} dataset was established for developing HAR algorithms using smartphones. The data came from 30 volunteers (of age 19-48) performing 6 different activities: walking, walking upstairs, walking downstairs, sitting, standing and laying down.
The volunteers followed a protocol which lasts 192 seconds, performing it twice, with a Samsung S II mounted on the left side of a belt around the waist for the first time, and the same phone placed on a location on the belt where the user preferred during the second time. Sensor signals from the accelerometer and the gyroscope were collected at 50 Hz.

\subsubsection{WISDM} The WISDM (Wireless Sensor Data Mining) project~\cite{wisdm} was an early study set to explore issues in obtaining sensor data from mobile devices. The data was collected from 29 volunteers performing 6 activities of daily living which are common in daily routines: walking, jogging, sitting, standing, walking upstairs and walking downstairs.
The participants carried a smartphone (Nexus One, HTC Hero, or Motorola Backflip) in their trousers' front pocket and performed a varying number of times for each activity. The data from the accelerometer embedded in the smartphone was collected at a sampling frequency of 20 Hz. An approval from the Fordham University IRB (Institutional Review Board) was obtained before the data collection.

\subsubsection{Fenland Study} 
The Fenland Study is a prospective cohort study of 12,435 men and women aged 35-65 with a primary objective of identifying behavioral, environmental, and genetic causes of obesity and type-2 diabetes. The study recruited participants from three different sites, excluding participants with clinically diagnosed diabetes mellitus, inability to walk unaided, terminal illness, clinically diagnosed psychotic disorder, pregnancy, or lactation. The Fenland study has been previously described in detail~\cite{fenland, fenland2}.
After a baseline clinic visit, a subsample of 2,100 participants was asked to wear a wrist accelerometer (GeneActiv) %
on the non-dominant wrist. This device recorded triaxial acceleration at 60 Hz. Participants were instructed to wear both waterproof monitors continuously for six full days and nights during free-living conditions, including during showering and while they were sleeping. This subsample of participants constitutes the sampling frame for the current analyses.
Further, we obtained physical activity energy expenditure (PAEE) from the triaxial accelerometry sensor using a method previously described and validated in a previous study~\cite{white2016estimation}. This information was then used to obtain time-series data of metabolic equivalents (METs) and label them into sedentary ($\leq$ 0.5 METs), light physical activity (LPA) (0.5-3 METs) and moderate to vigorous physical activity (MVPA) ($\geq$ 3 METs) following the conversion 1 standard MET is 71 J$\cdot$min$^{-1}\cdot$kg$^{-1}$.

\subsection{Data Preparation} %
Minimal pre-processing was applied to the raw sensor datasets. The data was first normalized by applying z-normalization using the mean and standard deviation of the training data for each of the sensor channels. Each dataset is then segmented into sliding windows with size \(400 \times 3\), where 400 is the number of time stamps and 3 being the number of channels of a triaxial accelerometer. The sliding windows share \(50\%\) overlap, where the next window has a starting-time shifted 200 timestamps to the future. Data from \(20\% - 25\%\) of users from each labeled dataset was kept unseen to the models and used as the test set in order to evaluate the generalizability of the models. \rev{There is no re-sampling performed to standardize the sampling frequencies of the datasets. The same configurations and pre-processing procedures are kept across different datasets because we want to highlight that the performance achieved is a result of the training paradigm but not because of the differences in configurations or pre-processing procedures. Furthermore, a similar pre-processing protocol was also adopted in previous work \cite{multi_self_har} and, to make for a fair comparison, we followed similar pre-processing steps.}

\subsection{Experimental Setup} 

\subsubsection{\rev{Standard Evaluation}}

To evaluate the training pipeline irrespective of the neural network architecture, and to draw direct comparisons with previous work, a simple model architecture, TPN~\cite{multi_self_har}, was adopted. The model consists of a temporal convolutional core with task-specific heads attached to it. The core consists of three temporal (1D) convolutional layers, with 32, 64 and 96 filters and a kernel size of 24, 16, and 8 respectively, all with stride 1. The ReLU activation function is used with L2 regularization with a factor of \(0.0001\) for the weights. Between every pair of layers, a drop-out layer with a rate of \(0.1\) is used. A global 1D maximum pooling layer is connected to the last convolutional layer, and this forms the convolutional core.

Depending on the setup, different task-specific heads are used. For transformation discrimination tasks, the last layer of the convolutional core is connected to a fully connected (dense) layer of 256 hidden units, activated by ReLU. Following this is a fully connected layer of 1 unit, activated by the sigmoid function, which forms the output layer of one transformation discrimination task. Binary cross-entropy is used as the loss function. For activity recognition, a fully connected layer of 1024 hidden units is connected to the convolutional core, and activated by ReLU. An output prediction layer with the number of units equal to the number of activities is then attached with full connections to the previous dense layer and activated by the softmax function with the categorical cross-entropy as the loss function. It is important to note that when a model is trained on a multi-task dataset, the weights of the convolutional core are shared among all tasks. The losses of different tasks are summed with equal weightings during multi-task training. The overall architecture follows the common setup of a multi-task convolutional neural network.

During training, the Adam optimizer~\cite{adam} is used with the decay rate for the first moment being 0.9 and decay rate for the second moment being 0.999 (default settings), and a learning rate of 0.0003. The model is trained for a total of 30 epochs with early stopping, where the model with the lowest validation loss is picked as the final model to mitigate over-fitting. A confidence score threshold \(C\) of 0.5 is used to filter self-labeled samples in order to ensure plurality (which reflects high confidence), and at most the top \(K=10000\) samples for each class are chosen, maximizing the size of the dataset.

During fine-tuning, the weights of the first two layers of the convolutional core are fixed (frozen), where only weights of the third convolutional layer and subsequent layers are tuned when training. This design is to allow the model to fine-tune to the final target dataset while retaining previously learned knowledge. This was shown to be the most effective setup in previous work~\cite{multi_self_har}.

\rev{\subsubsection{Linear Evaluation}
In addition to the standard evaluation protocol mentioned above, the linear evaluation protocol, which is commonly adopted in semi-supervised learning techniques for computer vision \cite{image_simclr, linear_zhang2016colorful, linear_oord2018representation, linear_bachman2019learning}, is also used in this study. Under this evaluation protocol, after the model is pre-trained, the convolutional core is completely frozen, and the output is directly connected to a fully-connected layer followed by the softmax function. No additional nonlinearity is introduced for the new layer, and the score of the resulting classifier is used as a proxy for the quality of representations extracted by the convolutional core.

Similar to the classification head in standard evaluation, the linear layer is trained on the categorical cross-entropy loss with the Adam optimizer with the same parameters. The weights for the layer are randomly drawn from a normal distribution with the mean being 0 and the standard deviation being 0.01.

\subsubsection{Other Baseline Algorithms}
Apart from deep-learning algorithms, two additional semi-supervised learning algorithms, \textit{En-Co-training} \cite{en_co_training_guan2007activity} and \textit{Sparse-Encoding} \cite{sparse_coding_bhattacharya2014using}, are used for baseline comparison in our study.

Following the standard protocols proposed in the original work, we used an ensemble of a decision tree, a Na\"ive Bayes classifier and a 3-nearest neighbor classifier for \textit{En-Co-training} \cite{en_co_training_guan2007activity} (see section \ref{subsection:related_work_semi_supervised} for descriptions). Eight statistical features: (1) mean, (2) correlation between axes, (3) interquartile range (4) mean absolute deviation, (5) root mean square, (6) standard deviation, (7) variance and (8) spectral energy for each axis, as suggested by Yang et al. \cite{statistical_features_yang2007activity}, were extracted for each window. The pool size is set to be one-tenth of the number of samples in the unlabeled dataset, and the models are trained for 20 iterations.

Similarly for \textit{Sparse-Encoding}, a dictionary of 512 basis vectors were learned from the unlabeled data during training \cite{sparse_coding_bhattacharya2014using}. The basis vectors are then separated into 52 clusters, and the lower 10-percentile of vectors in each of the clusters which have low empirical entropy are discarded. A support vector machine with the radial basis function kernel is then trained using grid-search cross-validation. 
}

\section{Evaluation and Discussion}
In this section, several evaluation setups were adopted to test the performance of the proposed method in different settings. An in-depth comparison against \rev{baseline algorithms} is given in Section~\ref{section:results_main}, with a set of ablation studies to evaluate the effectiveness of the combination of pre-training tasks given in Section~\ref{subsection:results_ablation}. Experiments on the effect of training with limited availability of data are reported in Section~\ref{subsection:results_limited_data}. A visualization of the extracted features is given in Section~\ref{subsection:results_t_sne}, followed by a case study on the effect of different distributions of unlabeled data on the PAAS dataset in Section~\ref{subsection:results_paas_case_study}. %

\subsection{\rev{Comparison against Baseline Algorithms}} \label{section:results_main}

\subsubsection{Overview} \label{subsubsection:results_main_overview}
We tested the performance of the models using training data from \(75\% - 80\%\) of users, with testing data coming from the remaining set of users.
The activities that the models are tested on are dataset dependent.
We first built the baseline for comparison using models trained in a fully supervised manner. \rev{Results from three other semi-supervised learning algorithms, \textit{En-Co-Training} \cite{en_co_training_guan2007activity}, \textit{Sparse-Encoding} \cite{sparse_coding_bhattacharya2014using}, and transformation discrimination pre-training \cite{multi_self_har} were also reported.} 
Weighted F1 scores were used as the main evaluation metric. Five independent runs with different model initialization were performed for each setting in order to mitigate the effect of model initialization on performance. \rev{The average and the $95\%$ bootstrap confidence level of the metrics were reported. They were obtained by re-sampling the test set for 1000 iterations and evaluating the models on them, and the $2.5\%$ and $97.5\%$ percentiles of the evaluation metrics from the models were used to form the confidence intervals.}

\rev{Two different datasets were used as unlabeled datasets: MobiAct and HHAR. The} set of experiments was performed with MobiAct as the unlabeled dataset, except in the case of evaluating on MobiAct itself, the HHAR dataset was used as the unlabeled dataset instead. 

\rev{
\begin{table}
    \caption{\rev{Comparison of classification performance between \textit{SelfHAR} and other HAR algorithms. Weighted F1 scores were used to benchmark seven different labeled HAR datasets, where \textit{SelfHAR} outperformed the other algorithms in almost all cases. *MobiAct was trained with HHAR as the unlabeled dataset}} %
    \label{table:evaluation_overivew}
    {\small

    \begin{center}
        \begin{tabularx}{\textwidth}{C|CCCCC}
Dataset & Fully Supervised (Baseline) & En-Co-Training \cite{en_co_training_guan2007activity} & Sparse-Coding \cite{sparse_coding_bhattacharya2014using} & Transformation Discrimination (TD) \cite{multi_self_har} & \textit{SelfHAR} \\
\hline
HHAR  & 0.7282 \space\space\space\space\space\space\space\space\space\space [0.7241, 0.7378] & 0.6505 \space\space\space\space\space\space\space\space\space\space [0.6377, 0.6641] ($-7.77\%$) & 0.4140 \space\space\space\space\space\space\space\space\space\space [0.3517, 0.5019] ($-31.42\%$) & \boldmath{}\textbf{0.7961 \space\space\space\space\space\space\space\space\space\space [0.7905, 0.8021] ($+6.79\%$)}\unboldmath{} & 0.7846 \space\space\space\space\space\space\space\space\space\space [0.7794, 0.7917] ($+5.64\%$) \\
\hline
Motion Sense & 0.9275 \space\space\space\space\space\space\space\space\space\space [0.9144, 0.9404] & 0.7080 \space\space\space\space\space\space\space\space\space\space [0.6765, 0.7395] ($-21.95\%$) & 0.8015 \space\space\space\space\space\space\space\space\space\space [0.7735, 0.8278] ($-12.60\%$) & 0.9295 \space\space\space\space\space\space\space\space\space\space [0.9173, 0.9420] ($+0.20\%$) & \boldmath{}\textbf{0.9631 \space\space\space\space\space\space\space\space\space\space [0.9530, 0.9726] ($+3.56\%$)}\unboldmath{} \\
\hline
MobiAct & 0.9098 \space\space\space\space\space\space\space\space\space\space [0.9038, 0.9140] & 0.8512 \space\space\space\space\space\space\space\space\space\space [0.8439, 0.8584] ($-5.86\%$) & 0.7848 \space\space\space\space\space\space\space\space\space\space [0.7616, 0.8097] ($-12.50\%$) & 0.8922 \space\space\space\space\space\space\space\space\space\space [0.8852, 0.8965] ($-1.76\%$) & \boldmath{}\textbf{0.9355 \space\space\space\space\space\space\space\space\space\space [0.9305, 0.9394] ($+2.57\%$)}\unboldmath{} \\
\hline
PAAS  & 0.6088 \space\space\space\space\space\space\space\space\space\space [0.5985, 0.6222] & 0.5977 \space\space\space\space\space\space\space\space\space\space [0.5811, 0.6138] ($-1.11\%$) & 0.5338 \space\space\space\space\space\space\space\space\space\space [0.5083, 0.5611] ($-7.50\%$) & 0.6851 \space\space\space\space\space\space\space\space\space\space [0.6741, 0.6967] ($+7.63\%$) & \boldmath{}\textbf{0.7022 \space\space\space\space\space\space\space\space\space\space [0.6903, 0.7125] ($+9.34\%$)}\unboldmath{} \\
\hline
UniMiB & 0.7432 \space\space\space\space\space\space\space\space\space\space [0.6914, 0.7927] & 0.6596 \space\space\space\space\space\space\space\space\space\space [0.6002, 0.7145] ($-8.36\%$) & 0.5085 \space\space\space\space\space\space\space\space\space\space [0.4290, 0.5847] ($-23.47\%$) & 0.8098 \space\space\space\space\space\space\space\space\space\space [0.7600, 0.8516] ($+6.66\%$) & \boldmath{}\textbf{0.8731 \space\space\space\space\space\space\space\space\space\space [0.8350, 0.9094] ($+12.99\%$)}\unboldmath{} \\
\hline
UCI HAR & 0.9051 \space\space\space\space\space\space\space\space\space\space [0.8826, 0.9253] & 0.7926 \space\space\space\space\space\space\space\space\space\space [0.7590, 0.8244] ($-11.25\%$) & 0.7620 \space\space\space\space\space\space\space\space\space\space [0.7260, 0.8005] ($-14.31\%$) & 0.9053 \space\space\space\space\space\space\space\space\space\space [0.8832, 0.9275] ($+0.02\%$) & \boldmath{}\textbf{0.9135 \space\space\space\space\space\space\space\space\space\space [0.8929, 0.9354] ($+0.84\%$)}\unboldmath{} \\
\hline
WISDM & 0.8906 \space\space\space\space\space\space\space\space\space\space [0.8732, 0.9083] & 0.6989 \space\space\space\space\space\space\space\space\space\space [0.6706, 0.7259] ($-19.17\%$) & 0.6406 \space\space\space\space\space\space\space\space\space\space [0.5954, 0.6784] ($-25.00\%$) & 0.8948 \space\space\space\space\space\space\space\space\space\space [0.8780, 0.9136] ($+0.42\%$) & \boldmath{}\textbf{0.9081 \space\space\space\space\space\space\space\space\space\space [0.8921, 0.9243] ($+1.75\%$)}\unboldmath{} \\

        \end{tabularx}
    \end{center}
    }
\end{table}

}

Table~\ref{table:evaluation_overivew} shows the mean \rev{and bootstrap confidence intervals of weighted F1 scores attained by our proposed pipeline and other training algorithms} on the seven target datasets (HHAR, MotionSense, MobiAct, PAAS, UniMiB SHAR, UCI HAR, and WISDM) over five independent runs. Further, as reported in this table, our proposed method \rev{consistently outperformed the fully supervised training baseline, with performance gain up to $12.99\%$ in weighted F1 score on the UniMiB SHAR dataset over five independent runs. The improvements are statistically significant in HHAR, MotionSense, MobiAct, PAAS, and UniMiB SHAR, where the 95\% confidence intervals have no overlap between SelfHAR and the fully supervised training baseline.}

\rev{When compared against other semi-supervised training algorithms, deep-learning algorithms consistently outperformed the others, with performances lower than the baseline reported by \textit{En-Co-Training} \cite{en_co_training_guan2007activity} and \textit{Sparse-Encoding} \cite{sparse_coding_bhattacharya2014using}. This validated the ability of deep-learning algorithms in recognizing complex patterns in high-dimensional data.

Our proposed pipeline out-performed the transformation discrimination algorithm in six out of seven datasets. Statistically significant improvements can be observed in some of the datasets, including MotionSense, MobiAct, PAAS, and UniMiB, in which there is very little to no overlap between the confidence intervals of \textit{SelfHAR} and the second-best algorithm. Smaller improvements over the baseline were reported when evaluated against HHAR, which could be attributed to the heterogeneous set of devices used to collect the data, where the teacher model is more prone to incorrectly label samples in the knowledge distillation step. 
}

\subsubsection{Activity-level comparison}  \label{subsubsection:results_main_confusion_matrices}
\begin{figure}
    \begin{center}
        \includegraphics[width=\textwidth]{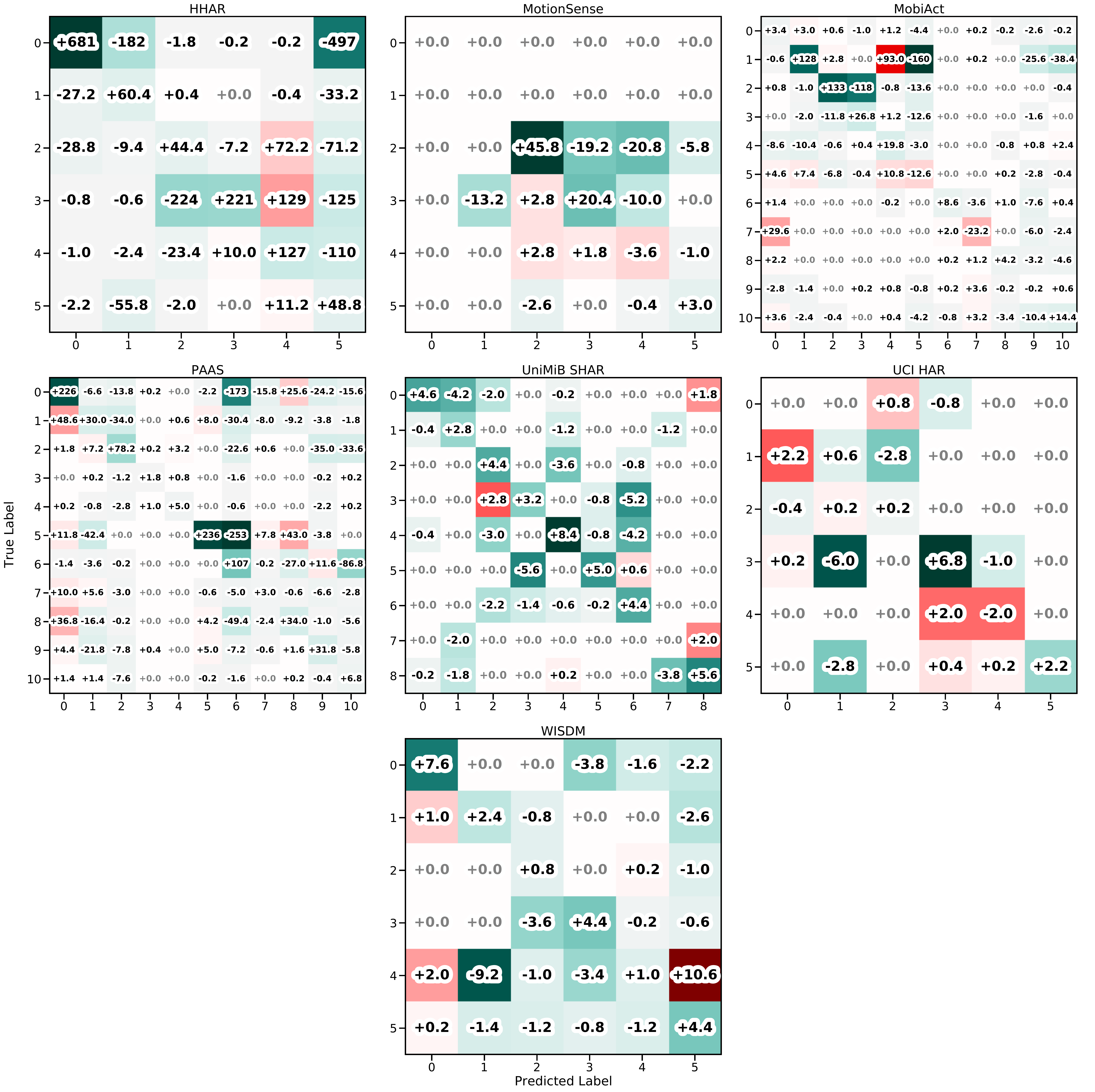}
    \caption[]
    {\rev{Comparison of model predictions at the activity (class) level. Delta of confusion matrices between \textit{SelfHAR} (with MobiAct/HHAR as the unlabeled dataset) and the fully supervised models on seven different datasets. Green cells denote an improvement over the fully supervised models, with an increase in correctly labeled samples or a decrease in model confusion. Red cells indicate a decrease in performance. See Appendix \ref{appendix:activity_labels} for the activity classes which correspond to the numbers shown along the axes of the confusion matrices. %
    }}
    \label{figure:delta_mobiact}
    \end{center}
\end{figure}

In order to give a more in-depth comparison between models on the activity level, the delta of confusion matrices, which are obtained by subtracting one confusion matrix from another, are provided in Figure~\ref{figure:delta_mobiact}. 

Figure~\ref{figure:delta_mobiact} shows the differences between confusion matrices obtained by using the supervised pipeline and \textit{SelfHAR}. The confusion matrices across five independent runs were averaged before computing the difference. These delta confusion matrices were obtained by subtracting the confusion matrices of the supervised method from those obtained using our proposed pipeline. Green cells denote an improvement over the fully supervised models, while red cells indicate an increase in model confusion. Along the main diagonal, positive values indicate that the model performed better, where more samples were categorized correctly, while positive values in off-diagonal cells indicate an increase in the number of samples misclassified by the models.

Our results showed that overall, \textit{SelfHAR} improved over supervised models in most of the activities across all datasets. Positive improvements or similar performance across all classes can be seen on the HHAR, PAAS, UniMiB SHAR and WISDM datasets, which vary in terms of ranges of activities and sensor placements.  \textit{SelfHAR} models perform better in the vast majority of classes in the remaining datasets. This indicated that there was a general improvement across nearly all classes, rather than the model specializing in specific classes and sacrificing performance in other classes. 

\subsection{Ablation Studies: Effectiveness of the Combined Pre-training Tasks}   \label{subsection:results_ablation}

\begin{table}
    \caption{\rev{Evaluation of the classification performance of different configurations of the proposed training pipeline when using MobiAct as the unlabeled dataset. \textit{SelfHAR} outperforms other configurations in most datasets (metric: weighted F1 score). *MobiAct was trained with HHAR as the unlabeled dataset}}
    \label{table:ablation_evaluation_mobiact}
    {
    \small

    \begin{center}
    
        \begin{tabularx}{\textwidth}{C|CCCCC}
Dataset (Evaluation Protocol) & Fully Supervised (Baseline) & Transformation Discrimination (TD)  & Self-Training  & Transformation Knowledge Distillation & SelfHAR \\
\hline \hline
HHAR (Standard) & 0.7282 \space\space\space\space\space\space\space\space\space\space [0.7241, 0.7378] & \boldmath{}\textbf{0.7961 \space\space\space\space\space\space\space\space\space\space [0.7905, 0.8021] ($+6.79\%$)}\unboldmath{} & 0.7220 \space\space\space\space\space\space\space\space\space\space [0.7152, 0.7288] ($-0.62\%$) & 0.7724 \space\space\space\space\space\space\space\space\space\space [0.7667, 0.7786] ($+4.42\%$) & 0.7846 \space\space\space\space\space\space\space\space\space\space [0.7794, 0.7917] ($+5.64\%$) \\
\hline
MotionSense (Standard) & 0.9275 \space\space\space\space\space\space\space\space\space\space [0.9144, 0.9404] & 0.9295 \space\space\space\space\space\space\space\space\space\space [0.9173, 0.9420] ($+0.20\%$) & 0.9208 \space\space\space\space\space\space\space\space\space\space [0.9085, 0.9347] ($-0.67\%$) & 0.9177 \space\space\space\space\space\space\space\space\space\space [0.9037, 0.9314] ($-0.98\%$) & \boldmath{}\textbf{0.9631 \space\space\space\space\space\space\space\space\space\space [0.9530, 0.9726] ($+3.56\%$)}\unboldmath{} \\
\hline
MobiAct (Standard) & 0.9098 \space\space\space\space\space\space\space\space\space\space [0.9038, 0.9140] & 0.8922 \space\space\space\space\space\space\space\space\space\space [0.8852, 0.8965] ($-1.76\%$) & 0.9162 \space\space\space\space\space\space\space\space\space\space [0.9109, 0.9208] ($+0.64\%$) & 0.9086 \space\space\space\space\space\space\space\space\space\space [0.9024, 0.9134] ($-0.12\%$) & \boldmath{}\textbf{0.9355 \space\space\space\space\space\space\space\space\space\space [0.9305, 0.9394] ($+2.57\%$)}\unboldmath{} \\
\hline
PAAS (Standard) & 0.6088 \space\space\space\space\space\space\space\space\space\space [0.5985, 0.6222] & 0.6851 \space\space\space\space\space\space\space\space\space\space [0.6741, 0.6967] ($+7.63\%$) & 0.6459 \space\space\space\space\space\space\space\space\space\space [0.6342, 0.6577] ($+3.71\%$) & 0.6651 \space\space\space\space\space\space\space\space\space\space [0.6537, 0.6764] ($+5.63\%$) & \boldmath{}\textbf{0.7022 \space\space\space\space\space\space\space\space\space\space [0.6903, 0.7125] ($+9.34\%$)}\unboldmath{} \\
\hline
UniMiB (Standard) & 0.7432 \space\space\space\space\space\space\space\space\space\space [0.6914, 0.7927] & 0.8098 \space\space\space\space\space\space\space\space\space\space [0.7600, 0.8516] ($+6.66\%$) & 0.8264 \space\space\space\space\space\space\space\space\space\space [0.7842, 0.8666] ($+8.32\%$) & 0.8588 \space\space\space\space\space\space\space\space\space\space [0.8213, 0.8965] ($+11.56\%$) & \boldmath{}\textbf{0.8731 \space\space\space\space\space\space\space\space\space\space [0.8350, 0.9094] ($+12.99\%$)}\unboldmath{} \\
\hline
UCI HAR (Standard) & 0.9051 \space\space\space\space\space\space\space\space\space\space [0.8826, 0.9253] & 0.9053 \space\space\space\space\space\space\space\space\space\space [0.8832, 0.9275] ($+0.02\%$) & 0.9079 \space\space\space\space\space\space\space\space\space\space [0.8926, 0.9342] ($+0.28\%$) & 0.9135 \space\space\space\space\space\space\space\space\space\space [0.8920, 0.9332] ($+0.84\%$) & \boldmath{}\textbf{0.9135 \space\space\space\space\space\space\space\space\space\space [0.8929, 0.9354] ($+0.84\%$)}\unboldmath{} \\
\hline
WISDM (Standard) & 0.8906 \space\space\space\space\space\space\space\space\space\space [0.8732, 0.9083] & 0.8948 \space\space\space\space\space\space\space\space\space\space [0.8780, 0.9136] ($+0.42\%$) & 0.9045 \space\space\space\space\space\space\space\space\space\space [0.8950, 0.9272] ($+1.39\%$) & 0.9036 \space\space\space\space\space\space\space\space\space\space [0.8877, 0.9194] ($+1.30\%$) & \boldmath{}\textbf{0.9081 \space\space\space\space\space\space\space\space\space\space [0.8921, 0.9243] ($+1.75\%$)}\unboldmath{} \\
\hline\hline
HHAR (Linear) & 0.7235 \space\space\space\space\space\space\space\space\space\space [0.7189, 0.7325] & \boldmath{}\textbf{0.7507 \space\space\space\space\space\space\space\space\space\space [0.7448, 0.7584] ($+2.72\%$)}\unboldmath{} & 0.7174 \space\space\space\space\space\space\space\space\space\space [0.7106, 0.7247] ($-0.61\%$) & 0.7373 \space\space\space\space\space\space\space\space\space\space [0.7310, 0.7443] ($+1.38\%$) & 0.7327 \space\space\space\space\space\space\space\space\space\space [0.7261, 0.7403] ($+0.92\%$) \\
\hline
Motion Sense (Linear) & 0.9224 \space\space\space\space\space\space\space\space\space\space [0.9083, 0.9355] & \boldmath{}\textbf{0.9468 \space\space\space\space\space\space\space\space\space\space [0.9359, 0.9581] ($+2.44\%$)}\unboldmath{} & 0.9238 \space\space\space\space\space\space\space\space\space\space [0.9094, 0.9380] ($+0.14\%$) & 0.9062 \space\space\space\space\space\space\space\space\space\space [0.8905, 0.9207] ($-1.62\%$) & 0.9276 \space\space\space\space\space\space\space\space\space\space [0.9151, 0.9403] ($+0.52\%$) \\
\hline
MobiAct (Linear) & 0.9008 \space\space\space\space\space\space\space\space\space\space [0.8954, 0.9055] & 0.8812 \space\space\space\space\space\space\space\space\space\space [0.8749, 0.8865] ($-1.96\%$) & 0.8944 \space\space\space\space\space\space\space\space\space\space [0.8885, 0.8995] ($-0.64\%$) & 0.8623 \space\space\space\space\space\space\space\space\space\space [0.8552, 0.8678] ($-3.85\%$) & \boldmath{}\textbf{0.9216 \space\space\space\space\space\space\space\space\space\space [0.9163, 0.9269] ($+2.08\%$)}\unboldmath{} \\
\hline
PAAS (Linear) & 0.6228 \space\space\space\space\space\space\space\space\space\space [0.6131, 0.6365] & 0.6815 \space\space\space\space\space\space\space\space\space\space [0.6702, 0.6939] ($+5.87\%$) & 0.6158 \space\space\space\space\space\space\space\space\space\space [0.6045, 0.6289] ($-0.70\%$) & 0.6374 \space\space\space\space\space\space\space\space\space\space [0.6261, 0.6506] ($+1.46\%$) & \boldmath{}\textbf{0.6895 \space\space\space\space\space\space\space\space\space\space [0.6778, 0.7017] ($+6.67\%$)}\unboldmath{} \\
\hline
UniMiB (Linear) & 0.5806 \space\space\space\space\space\space\space\space\space\space [0.5179, 0.6437] & 0.5739 \space\space\space\space\space\space\space\space\space\space [0.5148, 0.6418] ($-0.67\%$) & 0.5516 \space\space\space\space\space\space\space\space\space\space [0.4898, 0.6201] ($-2.90\%$) & 0.5445 \space\space\space\space\space\space\space\space\space\space [0.4833, 0.6117] ($-3.61\%$) & \boldmath{}\textbf{0.6219 \space\space\space\space\space\space\space\space\space\space [0.5624, 0.6837] ($+4.13\%$)}\unboldmath{} \\
\hline
UCI HAR (Linear) & \textbf{0.8759 \space\space\space\space\space\space\space\space\space\space [0.8513, 0.8999]} & 0.8577 \space\space\space\space\space\space\space\space\space\space [0.8315, 0.8820] ($-1.82\%$) & 0.8631 \space\space\space\space\space\space\space\space\space\space [0.8350, 0.8891] ($-1.28\%$) & 0.8371 \space\space\space\space\space\space\space\space\space\space [0.8108, 0.8671] ($-3.88\%$) & 0.8707 \space\space\space\space\space\space\space\space\space\space [0.8457, 0.8950] ($-0.52\%$) \\
\hline
WISDM (Linear) & 0.8379 \space\space\space\space\space\space\space\space\space\space [0.8148, 0.8615] & 0.8672 \space\space\space\space\space\space\space\space\space\space [0.8461, 0.8871] ($+2.93\%$) & 0.8107 \space\space\space\space\space\space\space\space\space\space [0.7865, 0.8373] ($-2.72\%$) & 0.8042 \space\space\space\space\space\space\space\space\space\space [0.7798, 0.8312] ($-3.37\%$) & \boldmath{}\textbf{0.8811 \space\space\space\space\space\space\space\space\space\space [0.8629, 0.9017] ($+4.32\%$)}\unboldmath{} \\

        \end{tabularx}

    \end{center}
    }
\end{table}

\begin{table}
    \caption{\rev{Evaluation of the classification performance of different configurations of the proposed training pipeline when using Fenland as the unlabeled dataset. \textit{SelfHAR} outperforms other configurations in most datasets (metric: weighted F1 score).}}
    \label{table:ablation_evaluation_fenland}
    {\small

    \begin{center}
    
        \begin{tabularx}{\textwidth}{C|CCCCC}
        
Dataset (Evaluation Protocol) & Fully Supervised (Baseline) & Transformation Discrimination (TD)  & Self-Training  & Transformation Knowledge Distillation & SelfHAR  \\
\hline \hline
HHAR (Standard) & 0.7282 \space\space\space\space\space\space\space\space\space\space [0.7241, 0.7378] & \boldmath{}\textbf{0.8074 \space\space\space\space\space\space\space\space\space\space [0.8028, 0.8141] ($+7.92\%$)}\unboldmath{} & 0.7416 \space\space\space\space\space\space\space\space\space\space [0.7356, 0.7484] ($+1.34\%$) & 0.7811 \space\space\space\space\space\space\space\space\space\space [0.7747, 0.7872] ($+5.29\%$) & 0.7772 \space\space\space\space\space\space\space\space\space\space [0.7713, 0.7839] ($+4.90\%$) \\
\hline
MotionSense (Standard) & 0.9275 \space\space\space\space\space\space\space\space\space\space [0.9144, 0.9404] & 0.9101 \space\space\space\space\space\space\space\space\space\space [0.8953, 0.9244] ($-1.74\%$) & 0.9062 \space\space\space\space\space\space\space\space\space\space [0.8924, 0.9197] ($-2.13\%$) & 0.9291 \space\space\space\space\space\space\space\space\space\space [0.9156, 0.9416] ($+0.16\%$) & \boldmath{}\textbf{0.9515 \space\space\space\space\space\space\space\space\space\space [0.9406, 0.9625] ($+2.40\%$)}\unboldmath{} \\
\hline
MobiAct (Standard) & 0.9098 \space\space\space\space\space\space\space\space\space\space [0.9038, 0.9140] & 0.9112 \space\space\space\space\space\space\space\space\space\space [0.9054, 0.9152] ($+0.14\%$) & 0.9042 \space\space\space\space\space\space\space\space\space\space [0.8981, 0.9085] ($-0.56\%$) & 0.9242 \space\space\space\space\space\space\space\space\space\space [0.9192, 0.9291] ($+1.44\%$) & \boldmath{}\textbf{0.9249 \space\space\space\space\space\space\space\space\space\space [0.9198, 0.9290] ($+1.51\%$)}\unboldmath{} \\
\hline
PAAS (Standard) & 0.6088 \space\space\space\space\space\space\space\space\space\space [0.5985, 0.6222] & 0.7036 \space\space\space\space\space\space\space\space\space\space [0.6922, 0.7146] ($+9.48\%$) & 0.6730 \space\space\space\space\space\space\space\space\space\space [0.6625, 0.6839] ($+6.42\%$) & 0.6636 \space\space\space\space\space\space\space\space\space\space [0.6523, 0.6755] ($+5.48\%$) & \boldmath{}\textbf{0.7049 \space\space\space\space\space\space\space\space\space\space [0.6931, 0.7161] ($+9.61\%$)}\unboldmath{} \\
\hline
UniMiB (Standard) & 0.7432 \space\space\space\space\space\space\space\space\space\space [0.6914, 0.7927] & 0.8086 \space\space\space\space\space\space\space\space\space\space [0.7581, 0.8502] ($+6.54\%$) & 0.8216 \space\space\space\space\space\space\space\space\space\space [0.7770, 0.8632] ($+7.84\%$) & 0.8627 \space\space\space\space\space\space\space\space\space\space [0.8191, 0.8986] ($+11.95\%$) & \boldmath{}\textbf{0.8481 \space\space\space\space\space\space\space\space\space\space [0.8025, 0.8883] ($+10.49\%$)}\unboldmath{} \\
\hline
UCI HAR (Standard) & 0.9051 \space\space\space\space\space\space\space\space\space\space [0.8826, 0.9253] & 0.9218 \space\space\space\space\space\space\space\space\space\space [0.9014, 0.9425] ($+1.67\%$) & 0.9128 \space\space\space\space\space\space\space\space\space\space [0.8914, 0.9316] ($+0.77\%$) & 0.9177 \space\space\space\space\space\space\space\space\space\space [0.8973, 0.9374] ($+1.26\%$) & \boldmath{}\textbf{0.9237 \space\space\space\space\space\space\space\space\space\space [0.9039, 0.9430] ($+1.86\%$)}\unboldmath{} \\
\hline
WISDM (Standard) & 0.8906 \space\space\space\space\space\space\space\space\space\space [0.8732, 0.9083] & 0.9061 \space\space\space\space\space\space\space\space\space\space [0.8907, 0.9228] ($+1.55\%$) & 0.9104 \space\space\space\space\space\space\space\space\space\space [0.8944, 0.9270] ($+1.98\%$) & \boldmath{}\textbf{0.9154 \space\space\space\space\space\space\space\space\space\space [0.8996, 0.9309] ($+2.48\%$)}\unboldmath{} & 0.9090 \space\space\space\space\space\space\space\space\space\space [0.8930, 0.9253] ($+1.84\%$) \\
\hline\hline
HHAR (Linear) & 0.7235 \space\space\space\space\space\space\space\space\space\space [0.7189, 0.7325] & \boldmath{}\textbf{0.7996 \space\space\space\space\space\space\space\space\space\space [0.7942, 0.8060] ($+7.61\%$)}\unboldmath{} & 0.7004 \space\space\space\space\space\space\space\space\space\space [0.6936, 0.7071] ($-2.31\%$) & 0.7257 \space\space\space\space\space\space\space\space\space\space [0.7192, 0.7324] ($+0.22\%$) & 0.7393 \space\space\space\space\space\space\space\space\space\space [0.7334, 0.7470] ($+1.58\%$) \\
\hline
Motion Sense (Linear) & 0.9224 \space\space\space\space\space\space\space\space\space\space [0.9083, 0.9355] & 0.9106 \space\space\space\space\space\space\space\space\space\space [0.8947, 0.9257] ($-1.18\%$) & 0.9056 \space\space\space\space\space\space\space\space\space\space [0.8906, 0.9209] ($-1.68\%$) & 0.9310 \space\space\space\space\space\space\space\space\space\space [0.9179, 0.9439] ($+0.86\%$) & \boldmath{}\textbf{0.9421 \space\space\space\space\space\space\space\space\space\space [0.9293, 0.9533] ($+1.97\%$)}\unboldmath{} \\
\hline
MobiAct (Linear) & 0.9008 \space\space\space\space\space\space\space\space\space\space [0.8954, 0.9055] & 0.9061 \space\space\space\space\space\space\space\space\space\space [0.9002, 0.9108] ($+0.53\%$) & 0.8986 \space\space\space\space\space\space\space\space\space\space [0.8928, 0.9032] ($-0.22\%$) & 0.9053 \space\space\space\space\space\space\space\space\space\space [0.8997, 0.9108] ($+0.45\%$) & \boldmath{}\textbf{0.9167 \space\space\space\space\space\space\space\space\space\space [0.9115, 0.9216] ($+1.59\%$)}\unboldmath{} \\
\hline
PAAS (Linear) & 0.6228 \space\space\space\space\space\space\space\space\space\space [0.6131, 0.6365] & 0.6664 \space\space\space\space\space\space\space\space\space\space [0.6551, 0.6790] ($+4.36\%$) & 0.6383 \space\space\space\space\space\space\space\space\space\space [0.6265, 0.6506] ($+1.55\%$) & 0.6636 \space\space\space\space\space\space\space\space\space\space [0.6514, 0.6757] ($+4.08\%$) & \boldmath{}\textbf{0.6869 \space\space\space\space\space\space\space\space\space\space [0.6750, 0.6992] ($+6.41\%$)}\unboldmath{} \\
\hline
UniMiB (Linear) & 0.5806 \space\space\space\space\space\space\space\space\space\space [0.5179, 0.6437] & 0.5363 \space\space\space\space\space\space\space\space\space\space [0.4770, 0.6047] ($-4.43\%$) & 0.5369 \space\space\space\space\space\space\space\space\space\space [0.4802, 0.6055] ($-4.37\%$) & 0.5205 \space\space\space\space\space\space\space\space\space\space [0.4591, 0.5873] ($-6.01\%$) & \boldmath{}\textbf{0.6072 \space\space\space\space\space\space\space\space\space\space [0.5474, 0.6735] ($+2.66\%$)}\unboldmath{} \\
\hline
UCI HAR (Linear) & \textbf{0.8759 \space\space\space\space\space\space\space\space\space\space [0.8513, 0.8999]} & 0.8211 \space\space\space\space\space\space\space\space\space\space [0.7906, 0.8552] ($-5.48\%$) & 0.8569 \space\space\space\space\space\space\space\space\space\space [0.8299, 0.8853] ($-1.90\%$) & 0.8359 \space\space\space\space\space\space\space\space\space\space [0.8067, 0.8661] ($-4.00\%$) & 0.8734 \space\space\space\space\space\space\space\space\space\space [0.8464, 0.8983] ($-0.25\%$) \\
\hline
WISDM (Linear) & 0.8379 \space\space\space\space\space\space\space\space\space\space [0.8148, 0.8615] & 0.7944 \space\space\space\space\space\space\space\space\space\space [0.7700, 0.8229] ($-4.35\%$) & 0.8210 \space\space\space\space\space\space\space\space\space\space [0.7964, 0.8478] ($-1.69\%$) & 0.7626 \space\space\space\space\space\space\space\space\space\space [0.7369, 0.7906] ($-7.53\%$) & \boldmath{}\textbf{0.8605 \space\space\space\space\space\space\space\space\space\space [0.8390, 0.8823] ($+2.26\%$)}\unboldmath{} \\

        \end{tabularx}
        
    \end{center}
    }
\end{table}

\rev{
As our proposed pipeline can be separated into different components, it is useful to evaluate different configurations of the \textit{SelfHAR} pipeline through ablation testing, and to understand whether the novel combination of multiple supervisory signals improves the ability of the models to extract useful features from the data.}

Comparisons were drawn among \rev{five} different training \rev{configurations of the training pipeline (see section \ref{subsection:expanded_training_pipeline}): fully supervised, transformation discrimination, self-training, transformation knowledge distillation and \textit{SelfHAR}}.

The final models of the \rev{five} training pipelines have the same computational complexity as they shared exactly the same neural network architecture and the number of weights.

\rev{
In Table~\ref{table:ablation_evaluation_mobiact} and \ref{table:ablation_evaluation_fenland}, we compare the performance of \textit{SelfHAR} to the other four configurations. Our proposed SelfHAR pipeline achieved the highest performance in the vast majority of cases, irrespective of the unlabeled dataset used. In linear evaluation, a similar trend is observed, and in some cases, only the \textit{SelfHAR} pipeline saw performance improvements compared to the baseline, where other configurations all performed worse (such as in MobiAct and UniMiB in Table \ref{table:ablation_evaluation_mobiact}). However, the performance gap is smaller compared to the full model evaluation. This could be due to the training setup of the \textit{SelfHAR} pipeline: in normal training, the last convolutional layer of the student model is fine-tuned together with the classification head. However, in linear evaluation, the entire network is frozen, preventing the last convolutional layer to be fine-tuned. This supported the design choice of fine-tuning the last layer during training.
}

This set of ablation experiments confirmed our hypothesis that an increase in both internal diversity (through transformation discrimination) and external diversity (through the use of unlabeled datasets and self-training) of training data results in a better semi-supervised learning approach than focusing only on one aspect or having a fully supervised approach. 

This also indicates the effectiveness of the multi-task training paradigm, where combining several approaches gives better performance than a single approach, and in some cases, more than the individual approaches added together. \textit{SelfHAR} outperformed \rev{other configurations of the pipeline, including the previous state-of-the-art method in self-supervised HAR (the transformation discrimination task, or Transformation Prediction Network as proposed in~\cite{multi_self_har}), providing support for the specific design of the pipeline.} 

In addition, we observed the following trade-off in the results: while the fully supervised pipeline requires the least amount of training and performs the worst \rev{in most cases}, \textit{SelfHAR} requires the most amount of training but performs the best. The comparison is drawn between final models with the same architecture and number of weights but from different training pipelines. Due to the limited computing resources on mobile devices and the general scarcity of well-curated labeled datasets, the current need for deep learning models on wearable/mobile devices tends to focus on high accuracy and low latency. Our approach helps in this direction by improving performance without increasing the model complexity at inference. %

\subsection{Impact of Training with Limited Labeled Data}  \label{subsection:results_limited_data}

In addition to training models with full availability of labeled data, the proposed method was also evaluated in scenarios where there are limited amounts of labeled data. This evaluation is designed to simulate scenarios where the resources for collecting labeled data are very limited, which might arise when a new set of sensors is introduced, or the set of target activities and sensor placements are changed. In this evaluation protocol, a fixed number of labeled samples per activity class were extracted from the labeled datasets, and they were the only labeled training data that the models were trained on. This procedure was used to evaluate the performance of the models when the availability of data is low.

For the fully supervised training pipeline, the models were trained on these extracted samples and evaluated directly. The transformation discrimination only approach follows closely to that when evaluated with full availability of labeled data (see section \ref{subsection:results_ablation}), where the models were pre-trained on the entire training set using the task of transformation discrimination (the activity labels were not used during pre-training), and then fine-tuned on the extracted samples. The \textit{SelfHAR} pipeline involves training a teacher model with the extracted samples, followed by the student model pre-trained with the labels generated by the teacher model and the signal transformations labels, on sensor data from both the labeled and unlabeled dataset. The student model was fine-tuned in a similar manner at the end with the extracted samples.

Either 2, 5, 10, 50 or 100 samples per activity class were extracted from the labeled dataset to simulate different degrees of availability of labeled data. Similar to other experiment protocols, five independent runs were performed for each setting.

\begin{figure}
    \begin{center}
        \includegraphics[width=\textwidth]{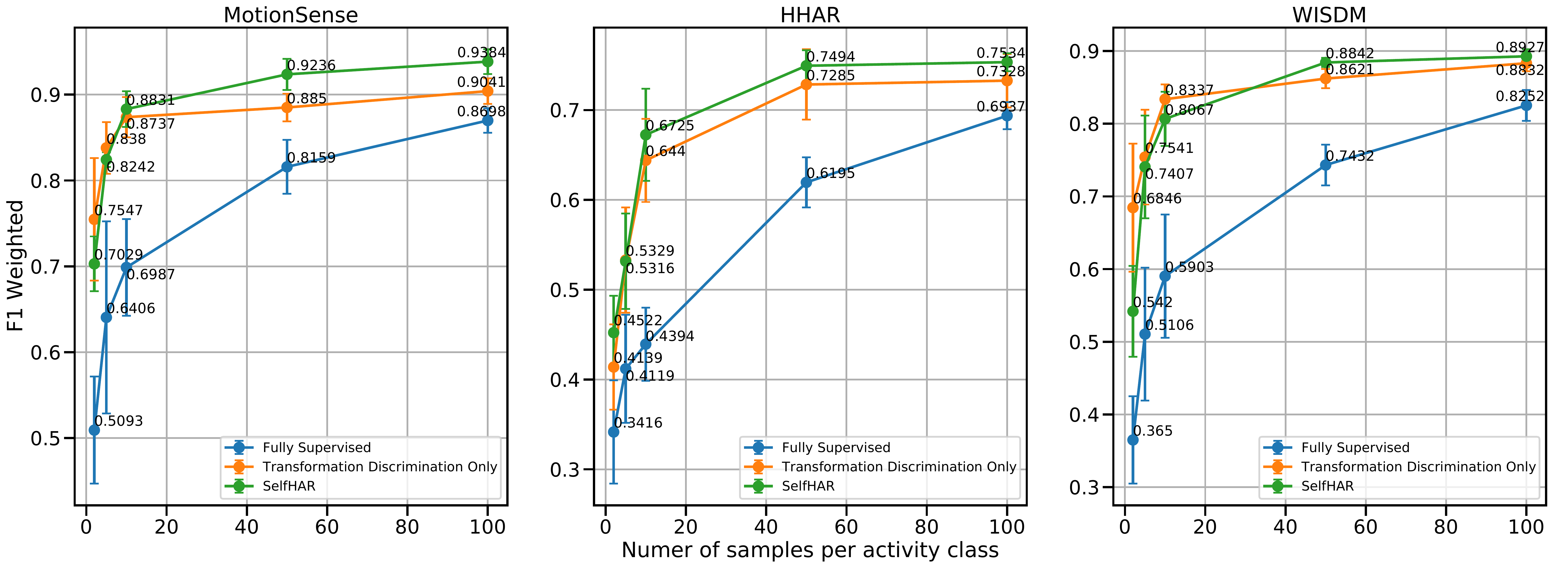}
    \caption{Assessing classification performance as a function of limited training data. \textit{SelfHAR} achieves high performance with significantly less training data and outperforms the variant with \textit{no} teacher-student training in most cases.}
    \label{figure:limited_data_plot}
    \end{center}
\end{figure}

Figure~\ref{figure:limited_data_plot} illustrates the performance of models trained by different pipelines with limited data availability. We observed significant performance differences between the \textit{SelfHAR} pipeline and the fully-supervised pipeline, especially when the amount of data available was particularly limited (2 - 10 samples per class). \rev{The difference in performance between the transformation discrimination only pipeline and SelfHAR varies depending on the datasets, with insignificant performance differences between them when there were only 2 - 10 samples per class, and the standard deviations are large.} However, the \textit{SelfHAR} pipeline showed a consistent performance improvement over other methods when there were 10 - 100 samples per class, with the performance gain being the largest in the MotionSense dataset.

The consistent performance gain across the range of data further indicated that the method can be adopted in general without penalty in performance. In cases of severe lack of data (labeled sample per class being less than 10), the transformation discrimination task on its own might perform better on some datasets. This could be attributed to the limitation of the teacher-student training paradigm: the uncertainty and noise in the supervisory signals could be amplified in the process when there are not many labeled samples to learn from. As the availability of data increases, the supervisory signals captured by the teacher-student training paradigm improve, and the \textit{SelfHAR} pipeline gives a higher performance gain than when using transformation discrimination alone.

\subsection{Visualization of Extracted Features Using t-SNE} \label{subsection:results_t_sne}

In addition to the \rev{confidence intervals} being reported, t-SNE visualizations~\cite{t_sne} of the features extracted by the last convolutional layer (after max pooling) of the compared models were also analyzed. t-Distributed Stochastic Neighbor Embedding (t-SNE) is an algorithm for projecting high-dimensional data points into two- or three- dimensional spaces, where the algorithm minimizes the difference between probability distributions of the high-dimensional and the low-dimensional space, giving representations that tend to cluster close-by similar data points~\cite{t_sne}. The minimization is done using a gradient-based method on the Kullback-Leibler divergence of the two probability distributions. The extracted features are visualized in order to give a better understanding of the models.

\begin{figure}
    \begin{center}
        \begin{tabularx}{\textwidth}{CC}
            (a) MotionSense & (b) UniMiB SHAR \\
            \includegraphics[width=0.5\textwidth]{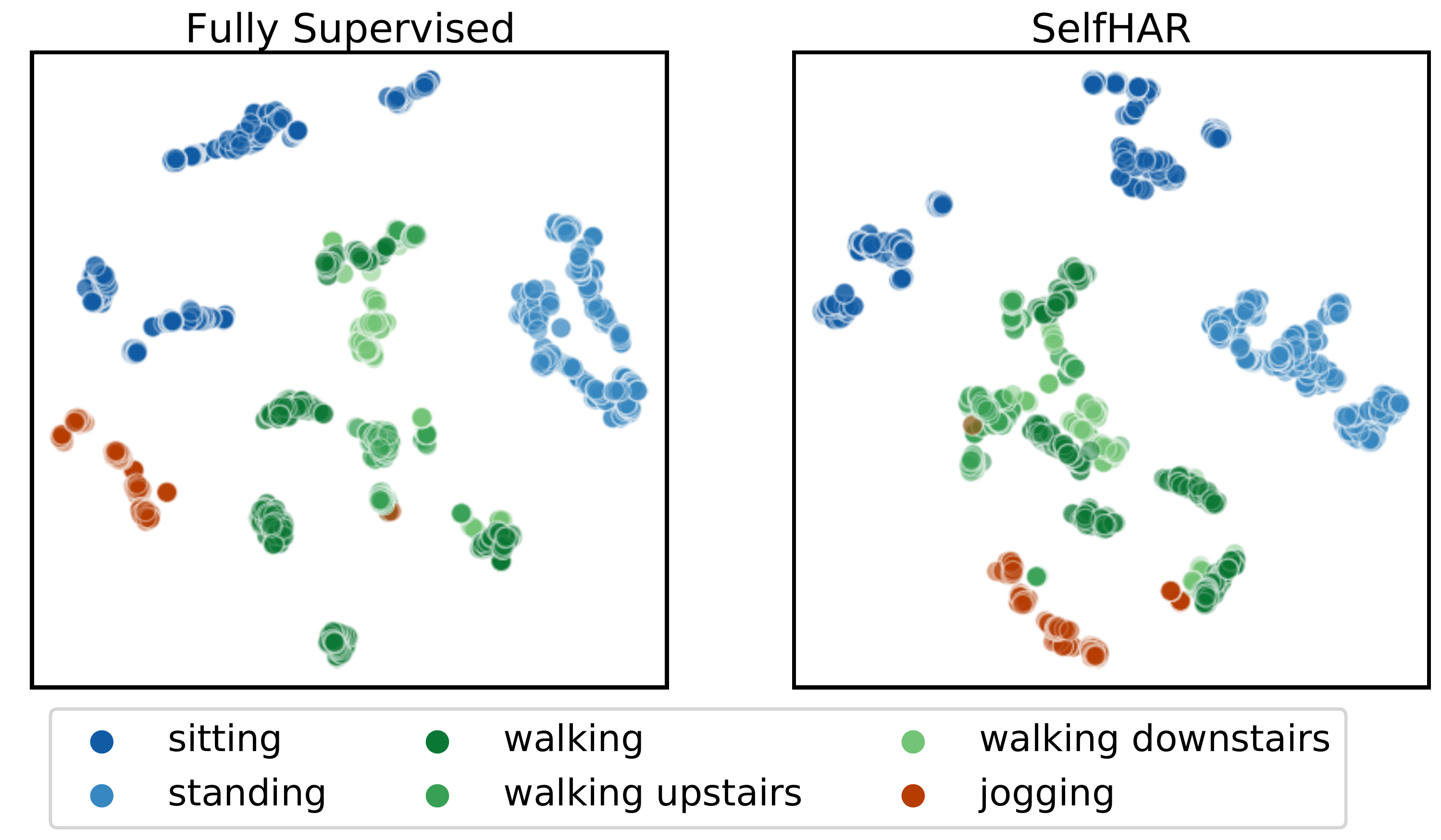} & \raisebox{-.19\height}{\includegraphics[width=0.5\textwidth]{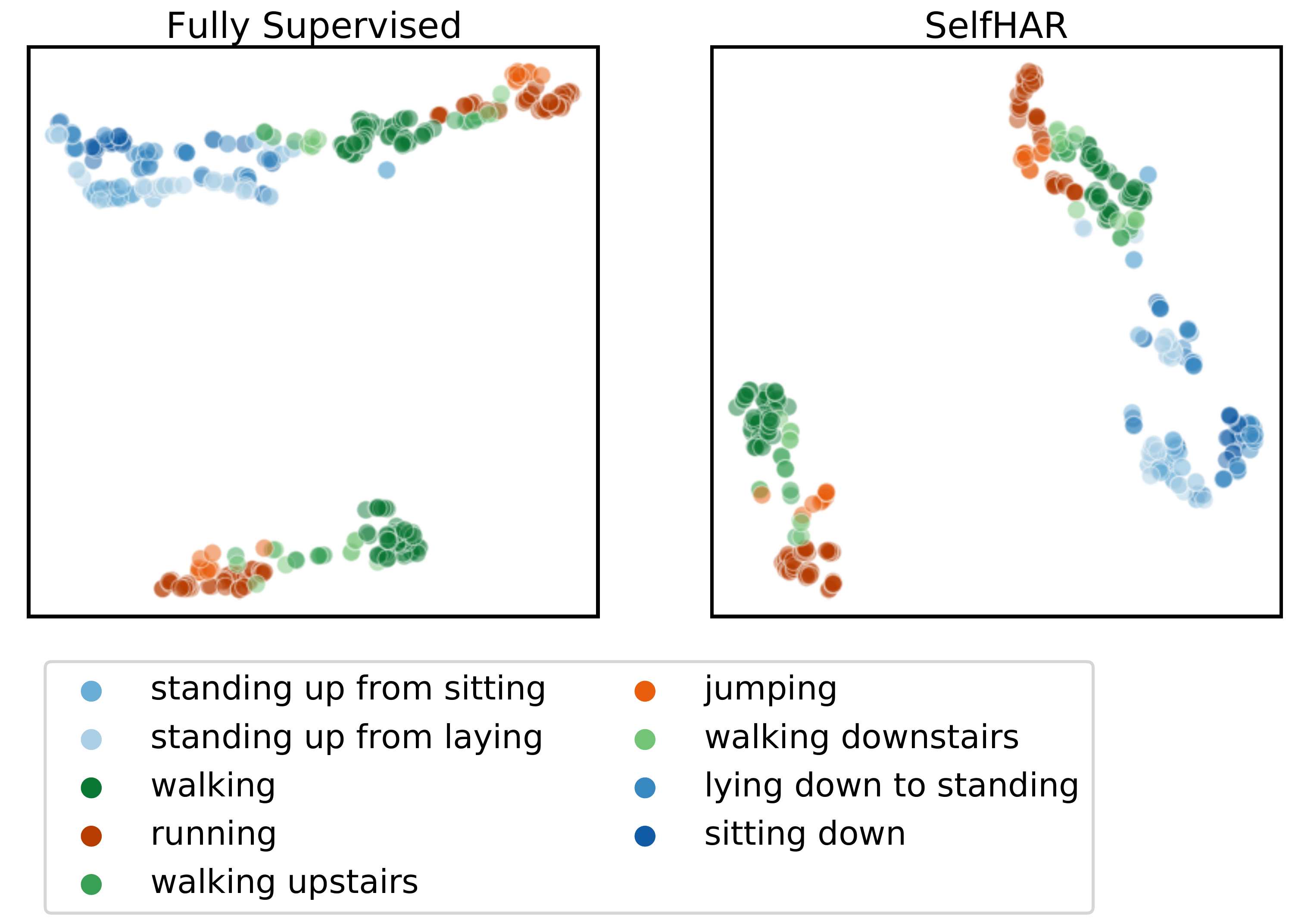}} \\
            (c) UCI HAR & (d) WISDM \\
            \includegraphics[width=0.5\textwidth]{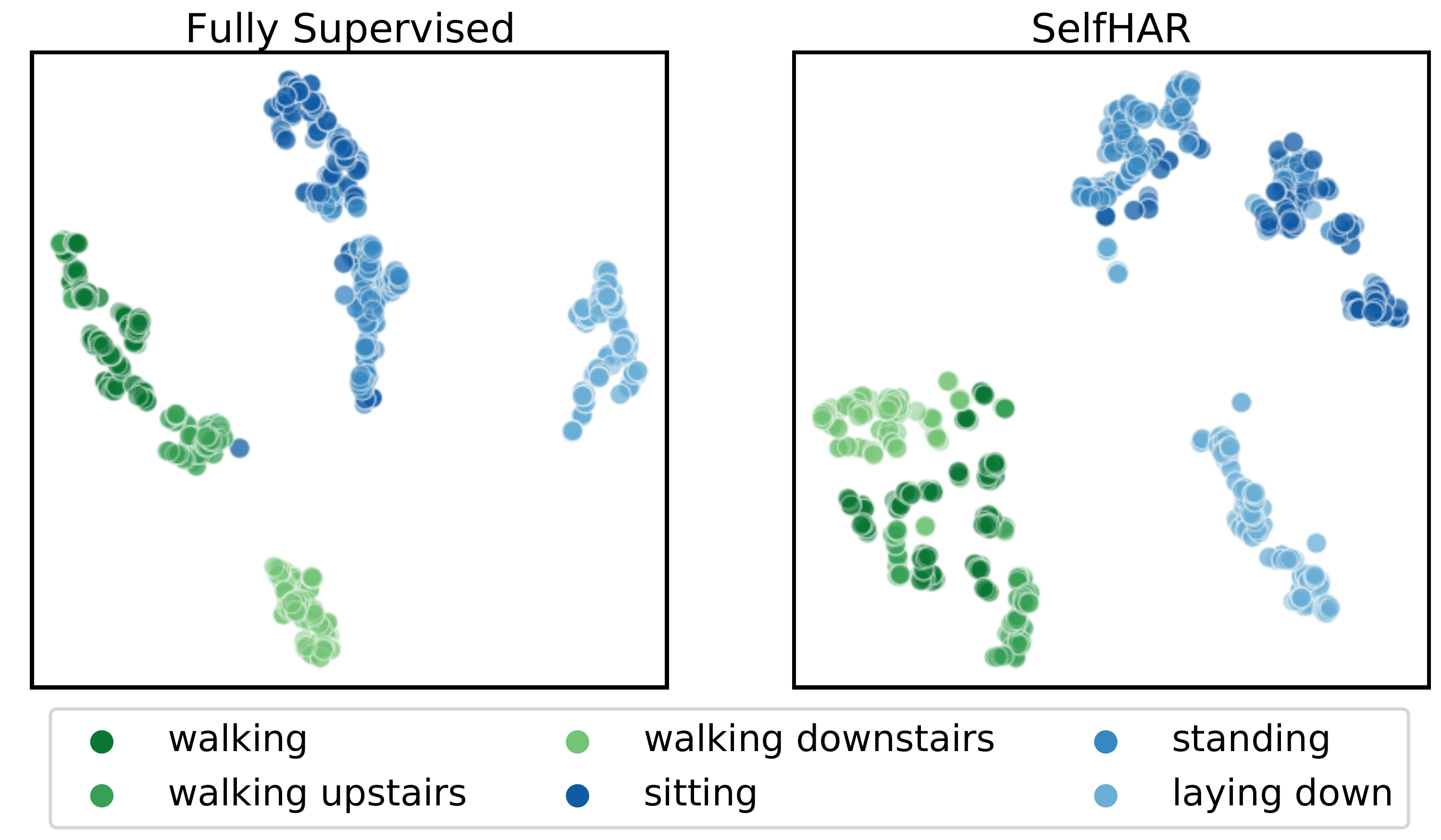} & \includegraphics[width=0.5\textwidth]{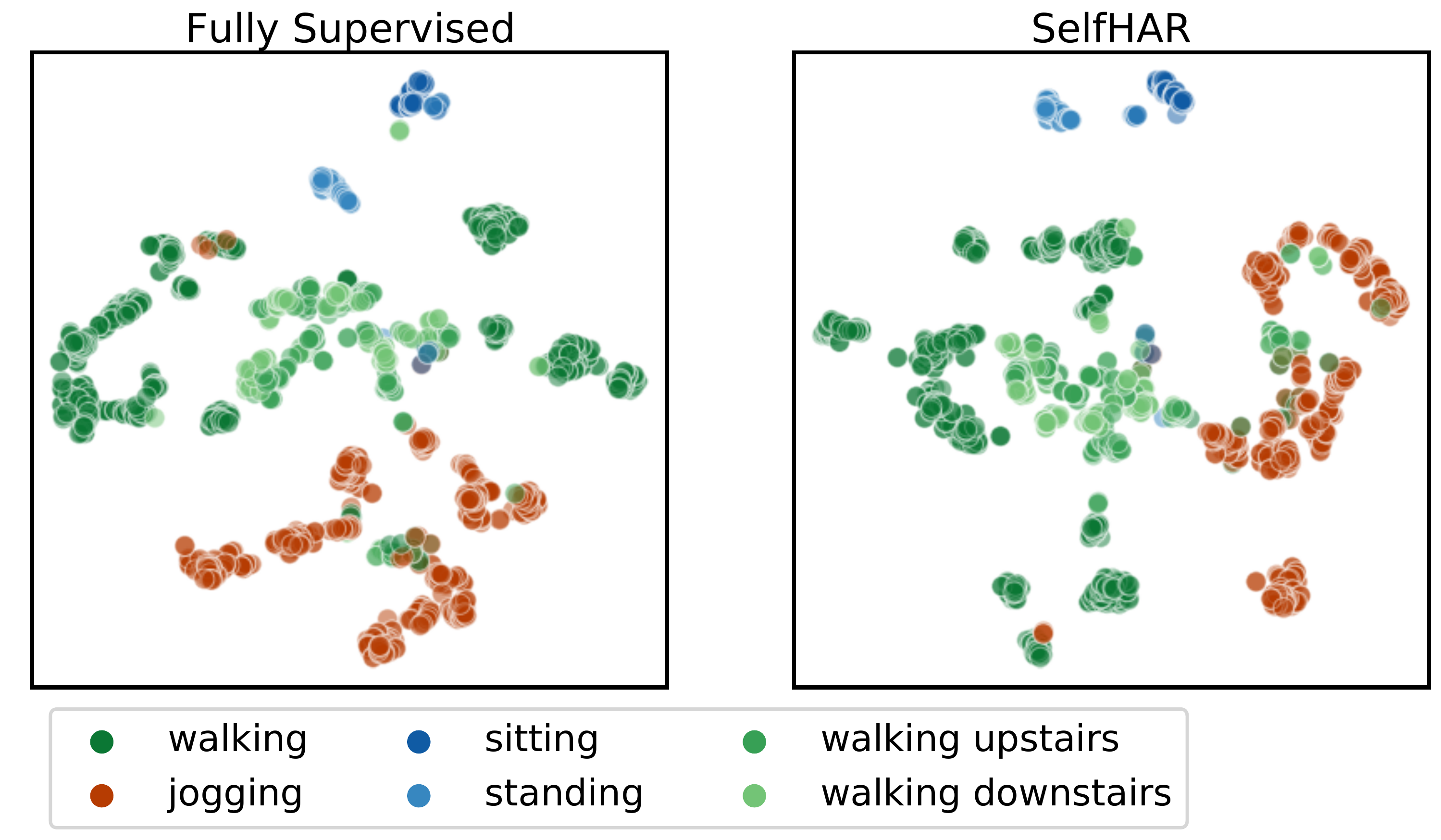} 
        \end{tabularx}
    \caption{Visualizing the learned representations of \textit{SelfHAR} versus the fully supervised model. T-SNE projections of the last convolutional layer showcase that the student before fine-tuning learns similar representations to the supervised model, even without direct access to the ground truth labels.} 
    \label{figure:t_SNE}
    \end{center}
\end{figure}

Figure~\ref{figure:t_SNE} displays the t-SNE projections of the extracted features by the models prior to fine-tuning, colored by their ground-truth classes (activity labels were not used when generating the t-SNE projection). The projections show that, without direct access to the ground truth labels, \textit{SelfHAR} discovered semantic manifolds within the data similar to the supervised model which had access to the labels. Similar results were obtained when using the rest of the datasets. We notice that the features largely correspond to those obtained with the supervised model. Notably, the clusters of data points across the two methods are similar in datasets such as WISDM, where the activity classes like jogging, walking and sitting were spread in a continuum according to their intensity.  %
These findings might imply that the performance boost did not come from over-fitting or mode collapse. It is evident that its inner workings resemble a supervised model.

\subsection{Leveraging large unlabeled datasets to improve supervised performance: PAAS and Fenland case study} \label{subsection:results_paas_case_study}

Upon designing our experiments, we hypothesized that the distribution of the dataset where pre-training took part may affect the results on the downstream task. Here, our aim was to test if having a similar distribution of activities or very different distributions would affect the results of the downstream task. Theoretically, through this process, the information captured by our architecture could be used to better discriminate in the downstream task.

Hence, for our evaluation, we constructed three different Fenland sub-datasets to test these effects using MET (metabolic equivalents) cutoffs. METs are a valuable tool for these cutoffs as they are well-established markers of energy expenditure. We decided to use Fenland and PAAS as both datasets used comparable devices placed on the same wrist.
The first dataset was constructed using only low MET values ($0.5 \leq MET \leq 1.5$, conveying mostly sedentary behaviors). The second one was a balanced dataset with the same number of samples from three MET distributions: low ($0.5 \leq MET \leq 1.5$), medium ($1.5 \leq MET \leq 3.0$) and high ($MET \geq 3.0$). Finally, an active dataset was built using only MET values $\geq 3.0$. All three datasets used a nearly identical number of total samples, allowing us to query the role of activity intensity distribution.

\begin{table}
    \caption{\rev{Evaluation of the classification performance using different filtering techniques for free-living accelerometer data. Pre-training \textit{SelfHAR} with balanced physical activity data outperforms the baseline and other variants.}}
    
    \label{table:fenland_paas}
    
    \begin{center}
        \begin{tabularx}{\textwidth}{C|CCCC}
    \multirow{2}{*}{Evaluation Metric}  & \multicolumn{4}{c}{Unlabeled Dataset for Pre-Training} \\
\cline{2-5}
 & None (Fully Supervised, Baseline) & Fenland (Inactive) & Fenland (Balanced) & Fenland (Active) \\
\hline
F1 Weighted & 0.6088 \space\space\space\space\space\space\space\space\space\space\space\space [0.5985, 0.6222] & 0.6832 \space\space\space\space\space\space\space\space\space\space\space\space [0.6722, 0.6948] ($+7.44\%$) & \boldmath{}\textbf{0.7049 \space\space\space\space\space\space\space\space\space\space\space\space [0.6931, 0.7161] ($+9.61\%$)}\unboldmath{} & 0.6662 \space\space\space\space\space\space\space\space\space\space\space\space [0.6557, 0.6786] ($+5.74\%$) \\
\hline
F1 Macro & 0.4264 \space\space\space\space\space\space\space\space\space\space\space\space [0.3987, 0.4645] & 0.5308 \space\space\space\space\space\space\space\space\space\space\space\space [0.4906, 0.5659] ($+10.44\%$) & \boldmath{}\textbf{0.5608 \space\space\space\space\space\space\space\space\space\space\space\space [0.5243, 0.5961] ($+13.44\%$)}\unboldmath{} & 0.5394 \space\space\space\space\space\space\space\space\space\space\space\space [0.4968, 0.5724] ($+11.30\%$) \\
\hline
Cohen's Kappa & 0.5182 \space\space\space\space\space\space\space\space\space\space\space\space [0.5047, 0.5323] & 0.6156 \space\space\space\space\space\space\space\space\space\space\space\space [0.6023, 0.6280] ($+9.74\%$) & \boldmath{}\textbf{0.6465 \space\space\space\space\space\space\space\space\space\space\space\space [0.6327, 0.6596] ($+12.83\%$)}\unboldmath{} & 0.6004 \space\space\space\space\space\space\space\space\space\space\space\space [0.5853, 0.6128] ($+8.22\%$) \\

        \end{tabularx}
    \end{center}
    
\end{table}

The results of this analysis can be found in Table~\ref{table:fenland_paas}.
We observed that self-supervised pre-training with the balanced dataset yielded the best performance, \rev{improving upon the fully supervised task by $13.44\%$ in F1 macro. On the other hand, the low MET dataset yielded a performance improvement of $10.44\%$ while the active dataset improvement was $11.30\%$ on average in F1 macro.} %
These results show that a more balanced distribution of data gave greater improvement compared to extreme distributions of data (either very active or non-active). This indicates that an unlabeled dataset which covers the full range of activities is more desirable as a pre-training task, which confirms our hypothesis regarding how this dataset is able to increase the diversity of data better than skewed datasets. Further case studies on other large-scale unlabeled datasets might provide more insights into the influence of the distribution of unlabeled data.

\section{Conclusion and Future Directions}
In light of the inherent limitations present in mobile sensing datasets and their gathering, we introduced \textit{SelfHAR}, a training pipeline that combines \rev{self-training and self-supervised learning techniques} to allow deep learning models to generalize to unseen scenarios by leveraging unlabeled data. The combined training pipeline increases both the internal and external diversity of data that models are trained on. Compared to previous approaches, our method was able to be able to further boost the performance of activity recognition models by leveraging the abundance of unlabeled data \rev{to complement the limited labeled data}. Evaluation indicates that the models trained using \textit{SelfHAR} \rev{outperform} other supervised and semi-supervised training approaches, both in terms of overall performance and on individual activities, without increasing the complexity of the models at inference time. %
Our proposed training paradigm and findings contribute to the development of well-performing deep learning models with limited labeled data, which has been an important challenge for the mobile and wearable sensing communities. Future work may explore similar approaches to label large unlabeled datasets, which would have important implications for epidemiological and public health research. 

This work indicates the potential of using a combined pre-training scheme in leveraging unlabeled data. An interesting extension of this work is to evaluate the representations learned in multi-device, multi-sensor settings, and on different mobile sensing tasks other than HAR. The ability to transfer to other tasks can also be an interesting way to validate whether the representations learned are generalizable. Other semi-supervised or self-supervised training techniques could potentially be incorporated into the framework to better leverage unlabeled data.

\begin{acks}
This work is partially supported by Nokia Bell Labs through their donation for the Centre of Mobile, Wearable Systems and Augmented Intelligence to the University of Cambridge. CI.T is additionally supported by the Doris Zimmern HKU-Cambridge Hughes Hall Scholarship and from the Higher Education Fund of the Government of Macao SAR, China. D.S is supported by the Embiricos Trust Scholarship of Jesus College Cambridge, and EPSRC through Grant DTP (EP/N509620/1). I.P is supported by GlaxoSmithKline and EPSRC through an iCase fellowship (17100053). The authors declare that they have no conflict of interest with respect to the publication of this work.
\end{acks} 

\bibliographystyle{ACM-Reference-Format}
\bibliography{sample-base}

\newpage
\appendix

\section{Algorithm of \textit{SelfHAR}} \label{appendix:algorithm}

\begin{algorithm}[H]

\SetAlgoLined
\SetKwInOut{Input}{Input}
\SetKwInOut{Output}{Output}
\Input{Labeled dataset \(D\), unlabeled dataset \(U\), Transformation functions \(T\), Activity classes \(A\), Number of teacher training epochs \(E_T\), Number of student pre-training epochs \(E_P\), Number of student fine-tuning epochs \(E_S\)}
\Output{A Trained HAR Model model\_student}

\begin{lstlisting}[language=Python, 
basicstyle=\ttfamily\footnotesize, 
commentstyle=\color{gray},
showspaces=false,
showstringspaces=false
]

# Teacher model training
model_teacher = new_har_model()
for epoch in [1 ... ET]:
    minibatch_train(model_teacher, dataset=D, loss=CrossEntropy(), 
                    optimizer=GradientDescent())

# Use teacher to label signals
W = combine_datasets(D.X, U)
W_labels = model_teacher.predict(W)
W_dataset = (W, W_labels)

# Select most confident samples from the dataset
S = []
for activity_class in A:
    W_sorted = sort(W_dataset, key = W_dataset.label.get_entry(activity_class))
    W_filtered = filter(W_sorted, condition = W_sorted.label.get_entry(activity_class) >= C)
    W_selected = W_filtered.get_first_n_entries(n = K)
    S.append(W_selected)

# Augment the data and construct a multi-task training dataset
D_prime = []
for (signals, har_label) in S:
    y = [0, 0, ..., 0]
    D_prime.append( (signals, har_label, copy(y) )
    for transformation in T:
        y[transformation] = 1 # To mark the transformation performed on the sample
        D_prime.append( (transformation(signals), har_label, copy(y) )
        y[transformation] = 0

# Student Training
model_student = new_multitask_model()
for epoch in [1 ... EP]:
    minibatch_train(model_student, dataset=D_prime, loss=CrossEntropy() + 
                    BinaryTransformationLoss(), optimizer=GradientDescent())

# Student Fine-tuning
for layer in model_student.get_core_layers():
    freeze_weights(layer)
for epoch in [1 ... ES]:
    minibatch_train(model_student, dataset=D, loss=CrossEntropy(), 
                    optimizer=GradientDescent())

\end{lstlisting}

 \caption{\revtwo{\textit{SelfHAR} - Combined semi-supervised learning}}
 \label{algorithm:pipeline}
\end{algorithm}

\section{Activity labels} \label{appendix:activity_labels}
The activity classes corresponding to the numbers shown along the axes of the delta confusion matrices in Figure \ref{figure:delta_mobiact} are: \textbf{HHAR} - 0: sitting, 1: standing, 2: walking, 3: walking upstairs, 4: walking downstairs, 5: biking, \textbf{MotionSense} - 0: sitting, 1: standing, 2: walking, 3: walking upstairs, 4: walking downstairs, 5: jogging, \textbf{MobiAct} - 0: standing, 1: walking, 2: jogging, 3: jumping, 4: walking upstairs, 5: walking downstairs, 6: standing to sitting on a chair, 7: sitting on a chair, 8: sitting to standing, 9: stepping into a car, 10: stepping out of a car, \textbf{PAAS} - 0: sitting, 1: standing, 2: walking, 3: walking upstairs, 4: walking downstairs, 5: lying, 6: cycling, 7: home activities, 8: office activities, 9: personal care, 10: shopping, \textbf{UCI HAR} - 0: walking, 1: walking upstairs, 2: walking downstairs, 3: sitting, 4: standing, 5: laying down, \textbf{UniMiB SHAR} - 0: standing up from sitting, 1: standing up from laying, 2: walking, 3: running, 4: walking upstairs, 5: jumping, 6: walking downstairs, 7: lying down to standing, 8: sitting down, \textbf{WISDM} - 0: walking, 1: jogging, 2: sitting, 3: standing, 4: walking upstairs, 5: walking downstairs.

\rev{
\section{Components of the SelfHAR Pipeline} \label{appendix:configurations_of_selfhar}

\begin{figure}
    \begin{center}
        \includegraphics[width=\textwidth]{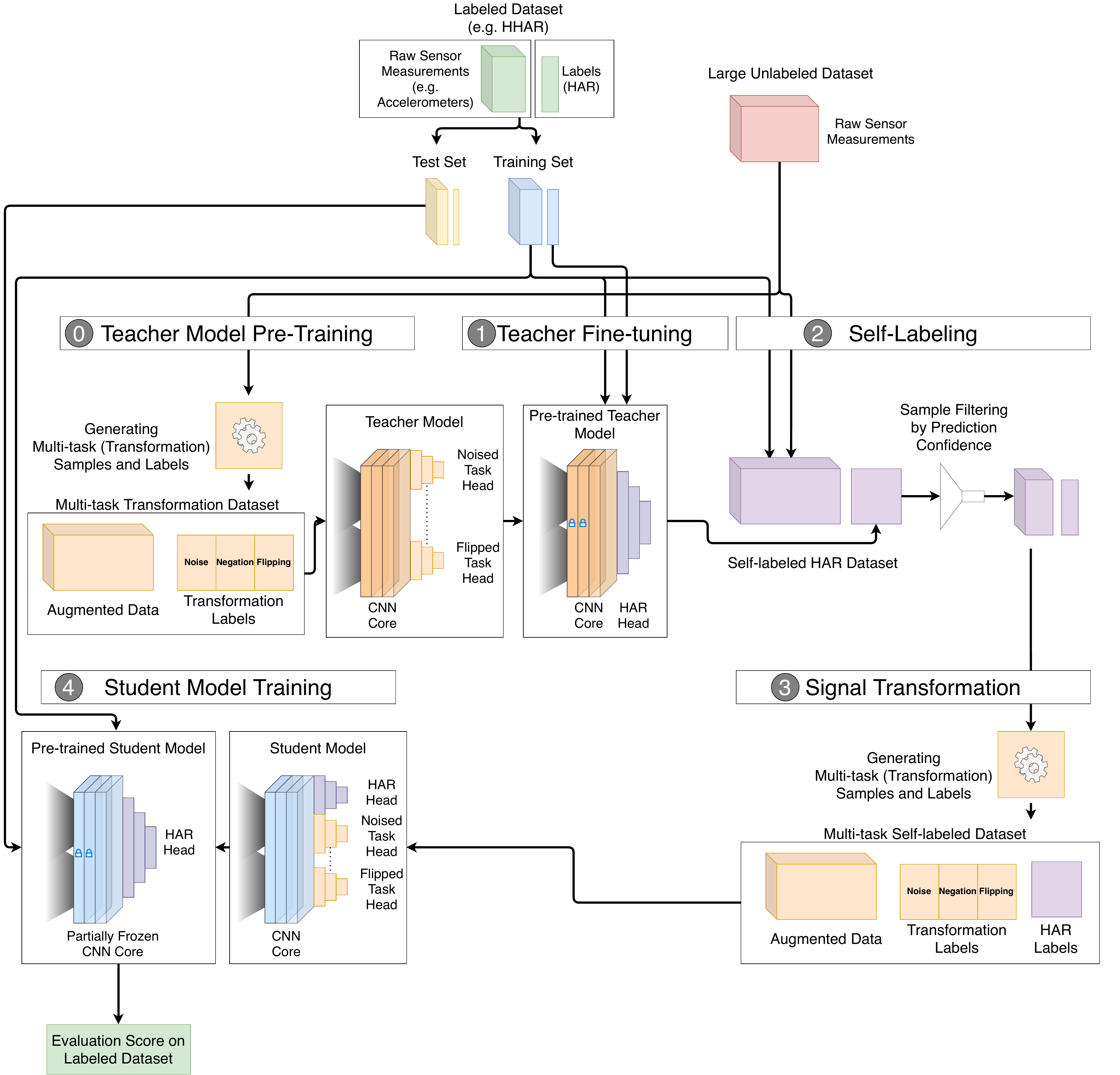}
    \caption{\rev{Components of the SelfHAR Pipeline. Five different configurations of the proposed pipeline were evaluated in our ablation studies: (1) fully supervised (using component 1 only), (2) transformation discrimination (using components 0 and 1), (3) self-training (using components 1, 2 and 4), (4) transformation knowledge distillation (using components 0, 1, 2 and 4) and (5) \textit{SelfHAR} (using components 1, 2, 3 and 4).}}
    \label{figure:architecture_expanded}
    \end{center}
\end{figure}

The five components that were used to form the different configurations of the \textit{SelfHAR} pipeline (as discussed in Section \ref{subsection:expanded_training_pipeline}) are visualized in Figure \ref{figure:architecture_expanded}.
} 

\end{document}